%% file: soccer.tex
\documentclass[sigconf]{acmart}
\AtBeginDocument{%
  }

\copyrightyear{2025}
\acmYear{2025}
\setcopyright{cc}
\setcctype{by}
\acmConference[KDD '25]{Proceedings of the 31st ACM SIGKDD Conference on
Knowledge Discovery and Data Mining V.2}{August 3--7, 2025}{Toronto, ON, Canada}
\acmBooktitle{Proceedings of the 31st ACM SIGKDD Conference on Knowledge
Discovery and Data Mining V.2 (KDD '25), August 3--7, 2025, Toronto, ON, Canada}
\acmDOI{10.1145/3711896.3737082}
\acmISBN{979-8-4007-1454-2/2025/08}

\acmSubmissionID{2181}

\usepackage{enumitem}
\usepackage{multirow}
\usepackage{makecell}
\usepackage{subcaption}

\begin{document}

\title{Player-Team Heterogeneous Interaction Graph Transformer for Soccer Outcome Prediction}

\author{Lintao Wang}
\affiliation{%
  \institution{The University of Sydney}
  \city{Sydney}
  \country{Australia}}
\email{lwan3720@uni.sydney.edu.au}

\author{Shiwen Xu}
\affiliation{%
  \institution{The University of Sydney}
  \city{Sydney}
  \country{Australia}}
\email{shiwen.xu0701@gmail.com}

\author{Michael Horton}
\affiliation{%
  \institution{Stats Perform Limited}
  \city{Wellington}
  \country{New Zealand}}
\email{michael.horton@statsperform.com}

\author{Joachim Gudmundsson}
\affiliation{%
  \institution{The University of Sydney}
  \city{Sydney}
  \country{Australia}}
\email{joachim.gudmundsson@sydney.edu.au}

\author{Zhiyong Wang}
\affiliation{%
  \institution{The University of Sydney}
  \city{Sydney}
  \country{Australia}}
\email{zhiyong.wang@sydney.edu.au}


\begin{abstract}
Predicting soccer match outcomes is a challenging task due to the inherently unpredictable nature of the game and the numerous dynamic factors influencing results.
While it conventionally relies on meticulous feature engineering, deep learning techniques have recently shown a great promise in learning effective player and team representations directly for soccer outcome prediction. 
However, existing methods often overlook the heterogeneous nature of interactions among players and teams, which is crucial for accurately modeling match dynamics. 
To address this gap, we propose \textbf{HIGFormer} (\textbf{H}eterogeneous \textbf{I}nteraction \textbf{G}raph \textbf{T}ransformer), a novel graph-augmented transformer-based deep learning model for soccer outcome prediction. 
HIGFormer introduces a multi-level interaction framework that captures both fine-grained player dynamics and high-level team interactions. Specifically, it comprises (1) a \textbf{Player Interaction Network}, which encodes player performance through heterogeneous interaction graphs, combining local graph convolutions with a global graph-augmented transformer; (2) a \textbf{Team Interaction Network}, which constructs interaction graphs from a team-to-team perspective to model historical match relationships; and (3) a \textbf{Match Comparison Transformer}, which jointly analyzes both team and player-level information to predict match outcomes.
Extensive experiments on the WyScout Open Access Dataset, a large-scale real-world soccer dataset, demonstrate that HIGFormer significantly outperforms existing methods in prediction accuracy. Furthermore, we provide valuable insights into leveraging our model for player performance evaluation, offering a new perspective on talent scouting and team strategy analysis.
\end{abstract}





\maketitle

%

\section{Introduction}

\begin{figure}[t] 
   \centering
   \includegraphics[width=\linewidth]{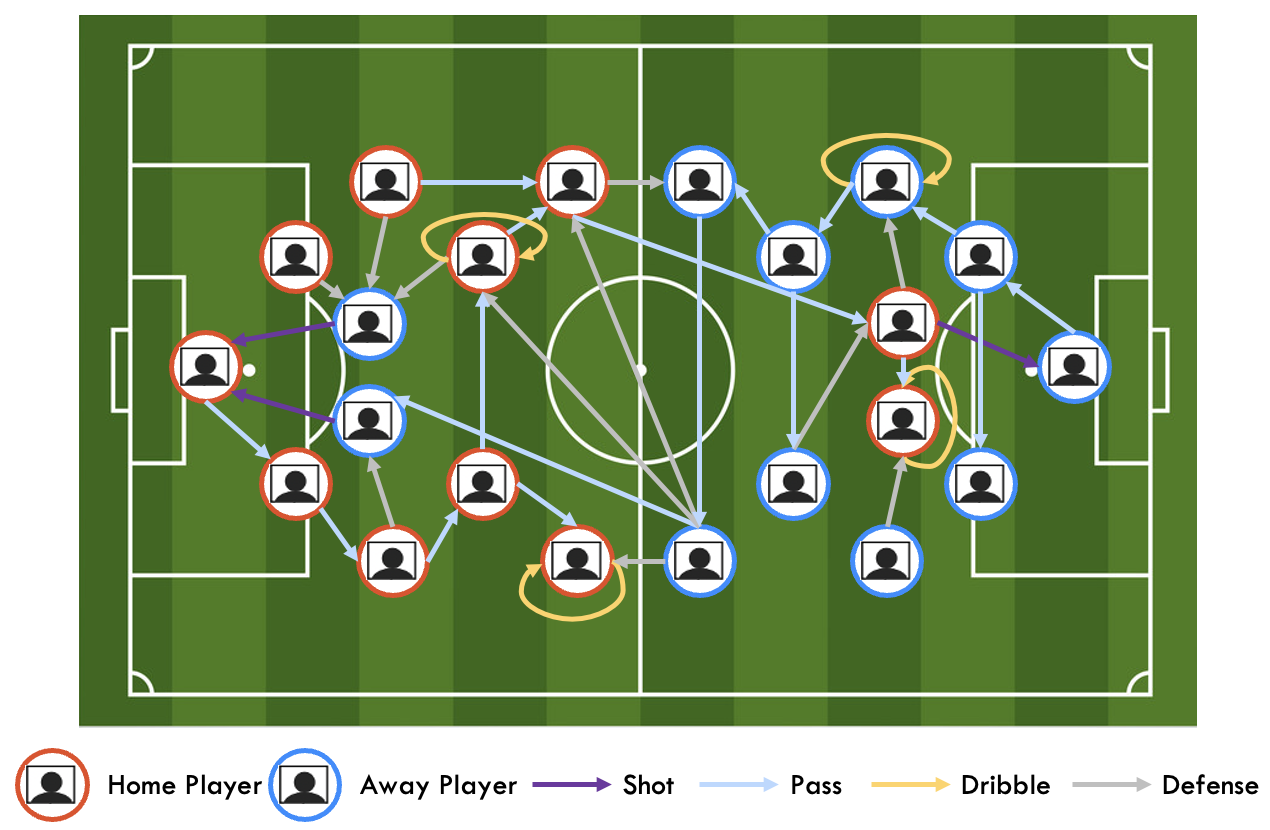}
   \caption{A soccer match can be modeled as an interaction graph for understanding match dynamics.}
   \label{fig:intro}
\end{figure}

Soccer (football) is the most popular sport in the world. Billions of fans are engaged, creating a market worth more than €29.5 billion ~\cite{market}. Thus, comprehensive soccer analysis provides tremendous value both strategically and financially. In data-driven analysis, soccer outcome prediction stands out as one of the most intuitive and compelling analyses, garnering significant interests from coaches, fans and others.
Nonetheless, predicting soccer outcomes presents challenges due to the unpredictable nature of the sport and many relevant factors affecting a match. 

Machine learning~\cite{2019domain,2019boost,2019bayesian} has been used to predict the outcomes of soccer matches. However, these methods frequently necessitate manual feature engineering by domain experts and are constrained by model capacity for complex predictive factors such as player interactions.  
Deep learning has emerged as a promising methodology for learning useful representations directly from data in various fields ~\cite{2016gcn,2017transformer}. While certain studies have investigated deep learning approaches for soccer outcome prediction ~\cite{soccer2020mlp, soccer2018rnn}, they are still limited by model capacity. This limitation stems from the fact that intrinsic feature learning heavily relies on input feature design rather than architectural design, as simple network structures are typically adopted.

It is considered that understanding player and team interactions is crucial for comprehending match dynamics and achieving accurate outcome predictions. This understanding can be mathematically simplified by representing a soccer game as an interaction graph (Figure ~\ref{fig:intro}). In this graph, players (nodes) from two teams (home team in red and away team in blue) interact with each other through various events (edges). Those events are typically discrete actions carried out by players such as passing and shooting.
For example, comprehending passing interactions between players can enhance our understanding of players' offensive performance and their impact on the final outcome. However, existing methods relying on hand-crafted features often fall short to model such complex heterogeneous relationship (various types of interactions).
To leverage such interaction information, there are two major challenges, namely 1) \textit{how to effectively formulate the interaction relationships}; and 2) \textit{how to efficiently learn from the interaction relationships}.

Therefore, to address the above-mentioned challenges, we propose HIGFormer (Heterogeneous Interaction Graph Transformer), a novel two-stream transformer-based framework that simultaneously captures historical player-player and team-team interaction. Unlike existing methods that overlook the heterogeneous nature of these interactions, HIGFormer explicitly models both fine-grained player dynamics and high-level team competition structures to enhance predictive accuracy.
For player-player interaction learning, we first construct a heterogeneous interaction graph for each historical match, where players serve as nodes and in-match events define the interaction edges. To extract meaningful heterogeneous representations, a Player Interaction Network is proposed, which integrates local graph convolution and global graph-augmented transformer. This dynamic combination enables effective learning from both localized and global contexts, leading to rich player embeddings that encode historical performance. 
For team-team interaction modeling, we represent team relationships through a win/loss interaction graph, where edges are based on historical winning rates. A Team Interaction Network applies graph convolution to derive team embeddings that encapsulate historical competition dynamics.
Finally, HIGFormer integrates both player and team embeddings using a Match Comparison Transformer, which aggregates historical interactions to predict match outcomes.
Comprehensive experiments on real-world soccer competitions demonstrate that HIGFormer significantly outperforms existing methods, highlighting its ability to effectively model multi-level soccer interactions for improved prediction accuracy. Furthermore, we explore its application in player evaluation, providing valuable insights that can support tactical decision-making in soccer.

The main contributions of this paper are as follows: 
\begin{itemize}
    \item We introduce HIGFormer, a novel Player-Team Heterogeneous Interaction Graph Transformer, which uniquely models multi-level player-team interactions to enhance match dynamics understanding and soccer outcome prediction.
    \item We propose a heterogeneous graph-based formulation for interaction modeling and develop a hybrid learning framework that combines local graph convolution with a global graph-augmented transformer, enabling more effective understanding of both fine-grained and high-level interactions.
    \item Comprehensive experimental results demonstrate the effectiveness of HIGFormer and highlight applications in player evaluation for valuable tactical decision-making insights.
\end{itemize}

\section{Related Work}
\subsection{Soccer Outcome Prediction}
Machine learning has been studied to predict soccer match outcomes by exploiting past match data. Early works ~\cite{soccer2006ma,soccer2013ma,soccer2015ma,soccer2016ma} applied various machine learning models with features constructed by domain experts. 
With the introduction of more comprehensive soccer datasets ~\cite{2019challenge,2019wyscout}, more complex machine learning models were proposed ~\cite{2019boost,2019domain,2019bayesian}. 
For example, \citet{2019boost} explored both relational and feature-based methods with features assessing historical performance such as Pi-Rating ~\cite{2013pirating} and PageRank ~\cite{2015pagerank}. 

Researchers also studied the application of artificial neural network for soccer outcome prediction. Multi-layer perceptron (MLP) ~\cite{soccer2010mlp} was applied to predict the outcome with eight simple features such as shots. \citet{soccer2011mlp} experimented more features including expert features as input to MLP and found that the expert features may not contribute to improvements. Techniques ~\cite{soccer2014mlp,soccer2017mlp,soccer2020mlp,soccer2022mlp} were further developed to apply MLP with different feature sets but few have made domain-specific improvements to the model structure. 
Additionally, recurrent neural network (RNN) was studied to model the historical match patterns for next match outcome prediction by either many-to-one ~\cite{soccer2018rnn} or many-to-many architecture ~\cite{soccer2020rnn}. It also enabled realtime match outcome prediction by dividing a match into interval segments and processing in a recurrent manner ~\cite{soccer2023rnn}. 

Nevertheless, manual feature engineering with machine learning methods often requires significant domain insights and faces challenges on modelling complex patterns. Furthermore, current deep learning methods are constrained by model capacity because intrinsic feature learning predominantly relies on input feature design rather than architectural design. By contrast, our method relies only on simple event features but exploits the intrinsic interaction patterns through architecture-based heterogeneous graph formulation and learning for soccer outcome prediction.

\subsection{Online Game Outcome Prediction}
Soccer match outcome prediction shares similarities with online game outcome prediction, where rich data availability has enabled various advanced methods. 
Early works ~\cite{2006bradley,2006trueskill} estimated team performance by aggregating individual scores but overlooked crucial player interactions. 
To address this, HOI ~\cite{2018hoi} exploited higher-order interactions using latent factor model, while OptMatch \cite{2020optmatch} and NeuralAC \cite{2021neuralac,2023massne} leveraged graph-based and attention-based mechanisms to capture cooperative and competitive effects. Transformer-based architecture ~\cite{2022draftrec} was also explored for modeling player histories and interactions with self-attentions.  
Beyond pre-match prediction, real-time outcome prediction has been studied using in-game features \cite{2019boostinggame,2022winningtracker}.

Similarly, our work emphasizes the crucial role of player interactions in outcome prediction. However, unlike prior methods tailored for online games, our approach introduces heterogeneous player interaction learning and team historical interaction modeling, specifically designed for soccer match analysis.

\subsection{Graph Learning}
Graph neural networks (GNN) have been extensively explored via spectral ~\cite{2016gcn,2018dualgcn,2019simplifygcn} or spatial domain ~\cite{2017graphsage,2017gat,2018gin} but limited in exploiting long-range relationships due to local message passing mechanism ~\cite{2019oversmoothing,2020oversquashing}.  
In contrast, transformers \cite{2017transformer} effectively capture global dependencies using self-attention, prompting efforts to integrate them with GNNs.
Some approaches combine local message passing with global attention ~\cite{2020grover,2020graphbert,2021graphtrans,2022gps,2024sgformer}. 
For example, GPS ~\cite{2022gps} formulated a general recipe to construct graph transformers by integrating the message passing neural network and the global attention layer to process input features. 
Others replace message passing entirely with pure attention mechanisms ~\cite{2020gt,2022tokengt,2022subgraphsat,2023goat,2023exphormer}, as seen in TokenGT ~\cite{2022tokengt}, which formulated Laplacian node identifiers and trainable type identifiers for input embeddings, performing attention on the entire graph to better capture global information. 

Real-world graphs are inherently heterogeneous, consisting of diverse node and edge types, leading to advances in heterogeneous graph learning \cite{2019hetergnn}. Methods either rely on predefined meta-paths ~\cite{2019han,2020magnn} or learn them adaptively ~\cite{2019gtn,2020hgt,2022mhgcn,2023hinormer}. While transformers have been explored for heterogeneous graphs ~\cite{2020hgt,2023hinormer}, they primarily focus on local structures rather than global relationships. 

Inspired by graph transformers’ success in global structure learning, we extend them into the heterogeneous context, incorporating heterogeneous GCNs for local information processing. To dynamically balance global and local insights, we introduce a novel mixture-of-experts (MoE) mechanism \cite{1991moe,2017moelstm,2022switchtransformer}, enabling adaptive fusion of these complementary representations.

\section{Methodology}

\begin{figure*}[h] 
   \centering
    \includegraphics[width=0.95\linewidth]{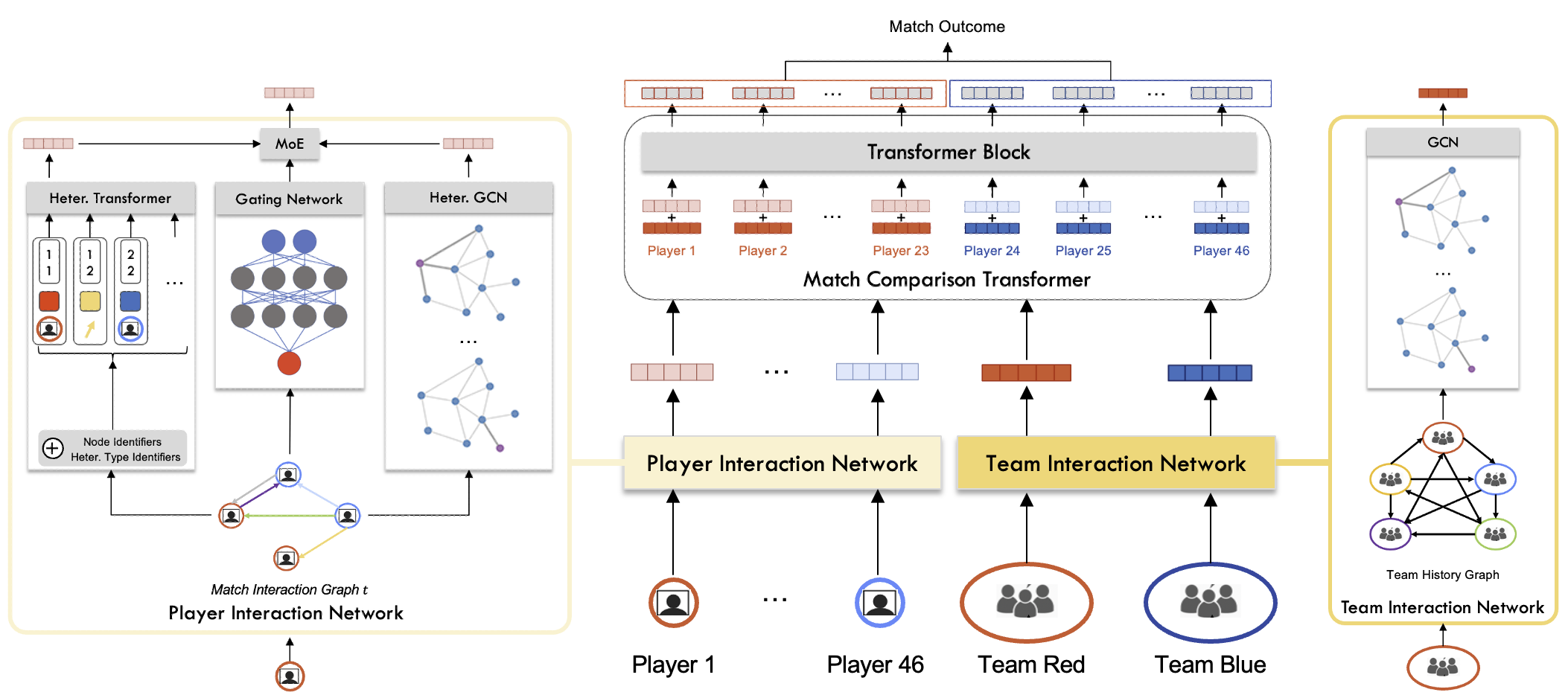}
   \caption{Illustration of HIGFormer, which consists of three major components: (1) Player Interaction Network for historical player performance learning, (2) Team Interaction Network for historical team performance comprehension and (3) Match Comparison Transformer for match outcome prediction.}
   \label{fig:arch}
\end{figure*}

Figure \ref{fig:arch} illustrates the proposed HIGFormer, which tackles the pre-match soccer outcome prediction problem (Section \ref{sec:preli}) with three major components. Player Interaction Network (Section \ref{sec:player}) encodes historical player performance into latent embedding by exploiting player-interaction-player based heterogeneous graph. Team Interaction Network (Section \ref{sec:team}) embeds historical team performance into latent space by utilizing winning rate based team interaction graphs. Match Comparison Transformer (Section \ref{sec:match}) ultimately predicts match outcomes by interpreting the performance embeddings of teams present in the match and their individual players. The entire model is trained in a two-stage manner for more efficient optimization (Section \ref{sec:loss}). 

\subsection{Problem Formulation} \label{sec:preli}
Our study centers on predicting pre-match outcomes for soccer matches, specifically utilizing historical match data. 
We denote the outcome of the $i$-th match as $ y^i \in \{\text{win, draw, lose}\}$.
Each match consists of two teams: Team Red $R^i$ and Team Blue $B^i$, each with 23 players including 11 starters. The players in Team Red are represented as $\{ p^i_{R,1}, \dots, p^i_{R,23} \}$ and those in Team Blue as $\{ p^i_{B,1}, \dots, p^i_{B,23} \}$. For simplicity, we denote any player in match $i$ as $p^i_n$, where $n$ refers to a player from either team.
Each player $p^i_n$ has a history of previous matches records as $\mathbf{H}_{p^i_n} = [\mathbf{h}^1_{p^i_n}, \ldots, \mathbf{h}^{i-1}_{p^i_n}]$. Similarly, the historical performance of a team is denoted as $\mathbf{H}_{R^i}$ and $\mathbf{H}_{B^i}$. 
To avoid extensive handcrafted feature engineering, we rely primarily on the historical records of players and teams as:
\begin{equation}
\begin{aligned}
    \mathbf{h}^{i-1}_{p^i_n} &= [\mathbf{c}^{i-1}_{p^i_n}, y^{i-1}_{p^i_n}], \\
    \mathbf{h}^{i-1}_{R^i} &= y^{i-1}_{R^i}, \\
    \mathbf{h}^{i-1}_{B^i} &= y^{i-1}_{B^i},  
\end{aligned}
\end{equation}
where $\mathbf{c}^{i-1}_p$ is the player's key event counts at past match $i-1$ and $y^{i-1}$ is the historical match $i-1$ outcome of a player or team as they may have different match histories. Specifically, a total of $10$ main events are selected, including {\it duel}, {\it foul}, {\it free kick}, {\it goalkeeper leaving the line}, {\it interruption}, {\it offside}, {\it others on the ball}, {\it pass}, {\it save attempt}, and {\it shot}.
The objective is to predict the match outcome $y^i$ using a learned function $f_{\theta}$ based on player and team histories: 
\begin{equation}
    \hat{y}^i = f_{\theta}\left(\left[\mathbf{H}_{p^i_n} | \text{for } n=1,\ldots,46\right], \mathbf{H}_{R^i}, \mathbf{H}_{B^i} \right).
\end{equation}

\subsection{Player Interaction Network} \label{sec:player}

\noindent\textbf{Heterogeneous Interaction Graph Formulation}. The interactions between players are vital for understanding their impacts to the final outcome. 
Instead of relying on handcrafted features which are often high-level and may overlook intrinsic complexity, we formulate the player interaction relationships into heterogeneous graphs to maintain its original diversity. For each match $i$, we define a heterogeneous player interaction graph as a directed graph: $G^i=(V^i, E^i, A, K)$ . 
Each player node $ v^i \in V^i $ corresponds to a player $p^i_n$ and is associated with a feature vector representing key event counts:
\begin{equation}
\mathbf{X}_V^i = [\mathbf{x}^{i}_{v,1}, \ldots, \mathbf{x}^{i}_{v,|V|}] = [\mathbf{c}^{i}_{v,i}, \ldots, \mathbf{c}^{i}_{v,|V|}],
\end{equation}
where $\mathbf{c}_{v,j}^i$ denotes the count of key events for player $v$ in match $i$. The node type mapping function is defined as $\tau(v): V \to A$, assigning each player to a category in $A$ which is Team Red and Blue in our case.
Each edge $ e^i \in E^i  $ depicts an interaction between players, characterized by an event type mapping function $\phi(e): E \to K$ and event-based features: 
\begin{equation}
\mathbf{X}_E^i = [\mathbf{x}^{i}_{e,k,1}, \ldots, \mathbf{x}^{i}_{e,k,|E|}]= [\mathbf{c}^{i}_{e,k,1}, \ldots, \mathbf{c}^{i}_{e,k,|E|}],
\end{equation}
where $\mathbf{c}_{e,k,j}^i$ denotes the count of event type $k$ between connected players in match $i$. In our problem, we establish two types of event edges which are pass-related events and defense-related events.

\noindent\textbf{Heterogeneous Graph Transformer (Heter. Transformer)}. To learn from heterogeneous graphs, we extend TokenGT \cite{2022tokengt}, which is originally designed for homogeneous graphs using a standard transformer architecture. The core idea is to formulate $|V|$ nodes and $|E|$ edges into a sequence of $|V|+|E|$ tokens as inputs to a standard transformer but with additional embedding based identifiers to incorporate graph inductive bias. 
We retain the Laplacian eigenvectors based node identifiers to indicate the graph structure, but extend them to integrate directional information. Additionally, we expand type identifiers to signify heterogeneous node and edge types, instead of distinguishing only nodes and edges. 
Specifically, we introduce three trainable embeddings: Node Type Embeddings $\mathbf{P}_{A}$, Edge Type Embeddings $\mathbf{P}_{K}$ and Node Identifiers $\mathbf{P}_{V}$. The feature augmentation process is defined as follows: 
\begin{itemize}
    \item Edge Feature Augmentation: For each edge $e^i \in E^i$ connecting node $u$ to $v$, we augment its features as $\mathbf{X}^i_{E,aug} = [\mathbf{X}^i_E, \mathbf{P}_u - \mathbf{P}_v + \mathbf{P}_k]$ where $k$ represents specific edge type and $\mathbf{P}_u - \mathbf{P}_v$ encodes a directional relationship from node $u$ to $v$.
    \item Node Feature Augmentation: For each node $v^i \in V^i$, its features are augmented as  $\mathbf{X}^i_{V,aug} = [\mathbf{X}^i_V, \mathbf{P}_v + \mathbf{P}_v + \mathbf{P}_a]$ where $a$ refers to specific node type and $\mathbf{P}_v + \mathbf{P}_v$ to align with directional relationship encoding for edge.  
\end{itemize}
The augmented token features $\mathbf{X}^i_{aug} = [\mathbf{X}^i_{V,aug}, \mathbf{X}^i_{E,aug}]$ serve as input to a transformer-based global player information encoder, denoted as $Enc^{glo}_{player}$: 
\begin{equation} \label{eqa:global}
\begin{aligned}
    \mathbf{Z}^{glo,i}_{player} &= Enc^{glo}_{player}\left(\{\mathbf{Q},\mathbf{K},V\} = f_{\{\mathbf{Q,K,V}\}}(\mathbf{X}^i_{aug})\right) \\
    &\in \mathbb{R}^{(|V|+|E|) \times d}, 
\end{aligned}
\end{equation}
where $\mathbf{Q}$, $\mathbf{K}$, $\mathbf{V}$ are Query, Key and Value in a transformer obtained by corresponding mapping function $f(\cdot)$. 

The output embedding $\mathbf{Z}^{glo,i}_{player} = [\mathbf{z}^{glo,i}_1,\ldots,\mathbf{z}^{glo,i}_{|V|+|E|}]$ captures global player and interaction information using long-range attention. As the edge embeddings are not utilized in subsequent operations, they are omitted from the final representations, yielding $\mathbf{Z}^{glo,i}_{player} \in \mathbb{R}^{|V| \times d}$. For simplicity, we retain the same notation for the final global player embeddings.

\noindent\textbf{Heterogeneous Graph Convolution Network (Heter. GCN)}.  As message-passing based graph neural networks still provide unique ability to capture local graph structure information ~\cite{2022gps,2023goat}, we formulate an additional GCN-based local branch.

Specifically, we adopt Graph Attention Network (GAT) ~\cite{2017gat} as the backbone. A GAT-based encoder $Enc^{loc}_{player}$ learns the local player interaction information as:
\begin{equation}\label{eqa:local}
    \mathbf{Z}^{loc,i}_{player} = Enc^{loc}_{player}\left(\mathbf{X}_{V}^i, \mathbf{X}_Enc^i\right) \in \mathbb{R}^{|V| \times d}.
\end{equation}
Since the original GAT is designed only for homogeneous graphs, we extend it to heterogeneous graphs by introducing separate trainable weights for different edge types $K$. The feature learning for node $v \in V$ at layer $l$ of heterogeneous GAT is given by:
\begin{equation}
\begin{aligned}
\mathbf{x}_v^{i,(l+1)\prime} & = \sum_{k=1}^{|K|} \sum_{u \in \mathcal{N}(v)} \alpha_{vu}^{l,k} \mathbf{W}^{l,k} \mathbf{x}_u^{i,(l)}, \\
\mathbf{x}_v^{i,(l+1)} & =\sigma\left(\mathbf{x}_v^{i,(l+1)\prime}\right),
\end{aligned}
\end{equation}
where $\mathcal{N}(v)$ is the local neighborhood of node $v$, $\alpha_{vu}^{l,k}$ is the attention score between nodes $v$ and $u$, $\mathbf{W}^{l,k}$ is the trainable weight matrix for edge type $k$ and $\sigma(\cdot)$ is an activation function. 
The attention score $\alpha^{l,k}_{vu}$ is computed using:
\begin{equation}
    \alpha_{vu} = \operatorname{Softmax}_{u \in \mathcal{N}(v)}\left(\mathbf{a}^\top[\mathbf{W}_{\alpha}\mathbf{x}_v, \mathbf{W}_{\alpha}\mathbf{x}_u] \right),
\end{equation}
where $\mathbf{a}$ is a learnable weight vector and $\mathbf{W}_{\alpha}$ is a learnable weight matrix. To retain the original node information, we add a self-loop for each node $v$.
The GAT-based encoder updates player embeddings by aggregating information from local neighbors using attention-weighted summation, effectively capturing local interactions.

\noindent\textbf{Mixture of Expert (MoE)}. To dynamically combine the local and global information, we adapt Mixture of Expert scheme ~\cite{1991moe} which aggregates the information from different experts based on the context. Compared to simple concatenation, MoE offers a more adaptive mechanism that enables representation learning to specialize on particular experts according to specific contextual cues. Specifically, there are two experts in our Player Interaction Network, namely global expert (Heter. Transformer) and local expert (Heter. GCN). A gating network, structured as MLP, predicts the expert weights $\mathbf{p}^i_{glo}$ and $\mathbf{p}^i_{loc}$ by leveraging player features $\mathbf{X}^i_{V}$ as: 
\begin{equation}
\begin{aligned}
    \relax[\mathbf{p}^i_{glo}, \mathbf{p}^i_{loc}] = \operatorname{Softmax}\left(\operatorname{Gating}(\mathbf{X}^i_{V})\right) \in \mathbb{R}^{|V| \times 2}, 
\end{aligned}
\end{equation}
where each player is associated with a unique local/global expert weight. 

The expert weights are subsequently utilized to calculate a weighted combination of local and global features, resulting in the final player embedding $\mathbf{Z}^{i}_{player}$ that harmonizes both the local and global information as:
\begin{equation}\label{eqa:moe}
    \mathbf{Z}^{i}_{player} = (\mathbf{p}^i_{glo} \odot \mathbf{Z}^{glo, i}_{player}) + (\mathbf{p}^i_{loc} \odot \mathbf{Z}^{loc, i}_{player}) \in \mathbb{R}^{|V| \times d}.
\end{equation}

\subsection{Team Interaction Network} \label{sec:team}
Soccer teams are persistent entities with additional functionalities such as youth systems and sports facilities, which can contribute to possible persistent dominance over others. To model this effect, we design Team Interaction Network, which captures team-level competition dynamics using a team interaction graph based on historical winning rates. We define the team interaction graph as a directed graph $G^{team} = (V^{team}, E^{team})$ where each node $v^{team} \in V^{team}$ represents a soccer team in competition and edge $e^{team} \in E^{team}$ represents the historical winning rate of one team against another. The node features are defined as learnable embeddings: $\mathbf{X}_V^{\text{team}} \in \mathbb{R}^{|V^{\text{team}}| \times d}$ and edge features are set as the historical winning rate: $\mathbf{X}_E^{\text{team}} \in \mathbb{R}^{|E^{\text{team}}|}$. 
We retain only the edge with the higher winning rate between two teams and set the direction accordingly. For example, if team A has won 20 out of 30 games against team B, we create a directed edge from A to B with a value of $\frac{2}{3}$.

Since teams consistently compete within the same competition, the team interaction graph is generally denser than the player interaction graph. Thus, we formulate a homogeneous GAT-based encoder \cite{2017gat}, denoted as $Enc_{team}$, to effectively capture local team interactions, which is sufficient to allow each team to attend to all other teams in the league. The encoded team representations are given by: 
\begin{equation}
    \mathbf{Z}_{team} = Enc_{team}\left(\mathbf{X}_V^{team}, \mathbf{X}_E^{team}\right) \in \mathbb{R}^{|V^{team}| \times d}.
\end{equation}

\subsection{Match Comparison Transformer} \label{sec:match}
By learning historical interactions for players and teams, we obtain embeddings $\mathbf{Z}_{player}$ and $\mathbf{Z}_{team}$ which capture rich historical performances information. 
To predict the match outcome, we introduce a Match Comparison Transformer $Enc_{match}$ then integrates and comprehends these embeddings. 

Since $\mathbf{Z}^i_{player}$ depends on match $i$ which is unknown before the match, we approximate them by computing historical pooled embeddings. Specifically, we construct $\mathbf{Z}^{i'}_{player}$ by applying average pooling over the last $T$ matches $(T \leq i - 1)$ for each player:
\begin{equation}
    \mathbf{Z}^{i'}_{player} = \left[[\operatorname{AvgPool}(\mathbf{z}^{t}_{p_n^i}|\text{for } t=i-1-T,\ldots,i-1)]\right],
\end{equation}
where $n=1,\ldots,46$ represents the 46 players participating in match $i$.
While the team embeddings $\mathbf{Z}_{team}$ are learned from the training set and do not depend on the test match $i$, we extract the embeddings for the two competing teams as $\mathbf{Z}^{i}_{team}$. These embeddings, fixed at test time, are added to player embeddings to identify each player’s associated team.

The Match Comparison Transformer $Enc_{match}$ processes the combined player and team embeddings as: 
\begin{equation}
\begin{aligned}
    \mathbf{Z}^{i}_{match} &= Enc_{match}\left(\{\mathbf{Q},\mathbf{K},V\} = f_{\{\mathbf{Q},\mathbf{K},V\}}(\mathbf{Z}^{i'}_{player} + \mathbf{Z}^{i}_{team})\right) \\
    &= [\mathbf{z}^{i}_{R,1},\ldots,\mathbf{z}^{i}_{R,23},\mathbf{z}^{i}_{B,1},\ldots,\mathbf{z}^{i}_{B,23}],
\end{aligned}
\end{equation}
where $\mathbf{z}^{i}$ represents updated embedding of a player from either Team $R$ (red) or Team $B$ (blue). The final match outcome $\hat{y}^i$ is predicted by comparing the aggregated team representations ~\cite{2022draftrec}, computed as the average player embeddings for each team: 
\begin{equation}
\begin{aligned}
    \mathbf{r}^i &= \operatorname{AvgPool} ([\mathbf{z}^{i}_{R,1},\ldots,\mathbf{z}^{i}_{R,23}]), \\
    \mathbf{b}^i &= \operatorname{AvgPool} ([\mathbf{z}^{i}_{B,1},\ldots,\mathbf{z}^{i}_{B,23}]), \\
    \hat{y}^i &= \sigma(f_{MLP}(\mathbf{r}^i - \mathbf{b}^i)), 
\end{aligned}
\end{equation}
where $f_{MLP}(\cdot)$ is a fully connected layer and $\sigma(\cdot)$ is a sigmoid activation function. The final prediction $\hat{y}^i$ represents the probability of Team $R$ winning.

\subsection{Model Training} \label{sec:loss}
Instead of treating match outcome prediction as a standard three-class classification problem with multi-class cross-entropy loss, we follow prior work \citep{soccer2018rnn, 2019boost} and apply Mean Squared Error (MSE) loss directly on the model output. This enables ordinal modeling of outcomes (Lose < Draw < Win), capturing the inherent relationships among classes more effectively. Specifically, given predicted match outcome $\hat{y}^i$, the ground truth outcome value $y^i$ is set as single value $1$ (win), $0.5$ (draw) and $0$ (lose) based on outcome type. The loss $\mathcal{L}_{out}$ is then defined as: 
\begin{equation}
    \mathcal{L}_{out} (y^i, \hat{y}^i)= \left\| y^i - \hat{y}^i \right\|^2.
\end{equation}

\begin{figure}[t] 
   \centering
   \includegraphics[width=\linewidth]{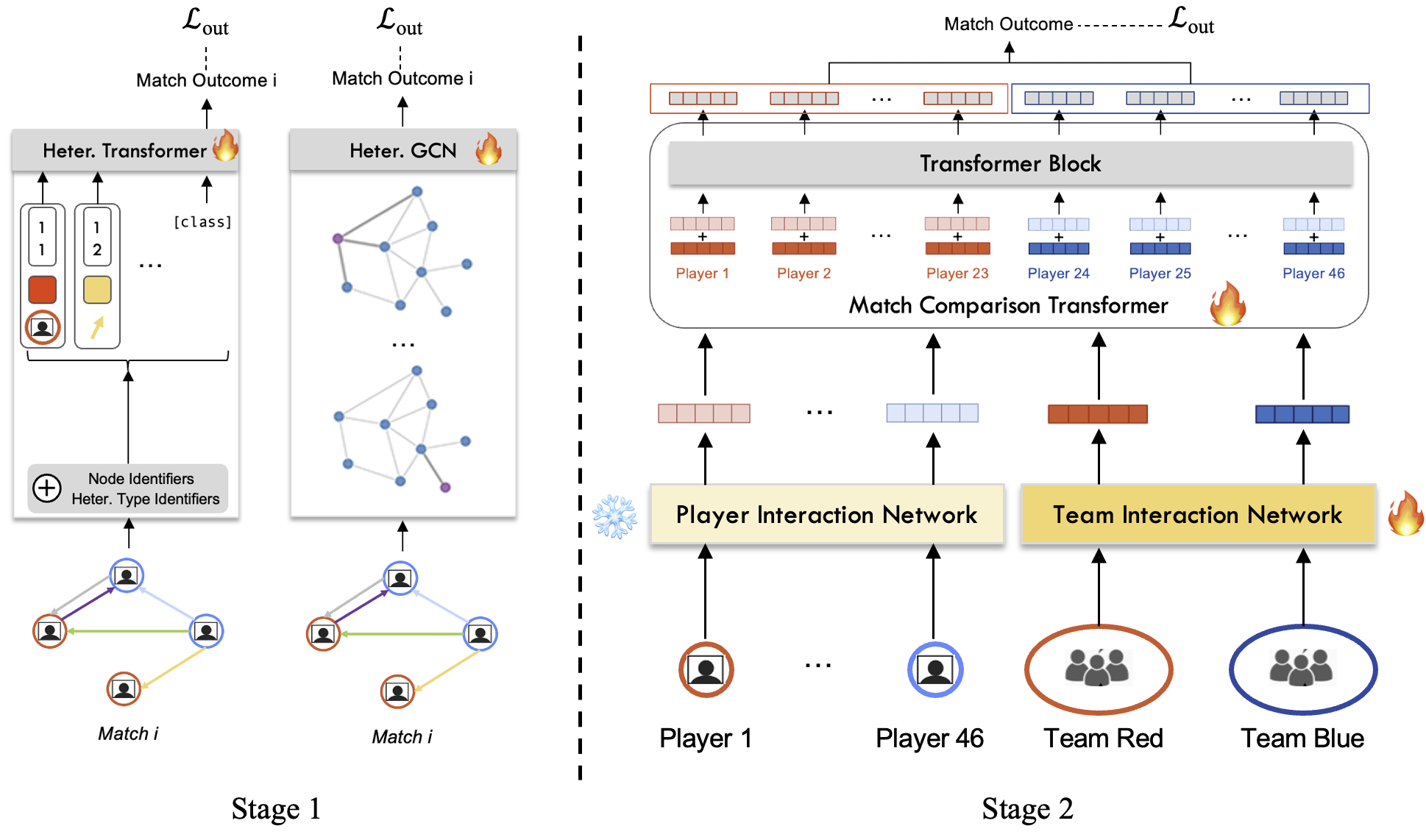}
   \caption{HIGFormer two-stage training where fire icon indicates trainable models and ice icon indicates frozen models. }
   \label{fig:train}
\end{figure}

Graph representation learning often relies on unsupervised or semi-supervised learning ~\cite{2018networksurvey}, as detailed labels for individual nodes in a graph are typically unavailable. In our task, the only available label is the graph-level match outcome. Since match outcome prediction involves aggregating information from multiple past matches (graphs), direct training is challenging due to extremely sparse supervision (a single label for multiple graphs). To address this, we design a two-stage training pipeline, as shown in Figure \ref{fig:train}.

In Stage 1, we pre-train the Heter. Transformer and Heter. GCN of the Player Interaction Network separately to characterize the player information per match. Instead of using historical player embeddings  $\mathbf{Z}^{i{\prime}}_{\text{player}}$, we construct a player interaction graph $G^i$ per match and use corresponding match player embeddings $\mathbf{Z}^{i}_{\text{player}}$. 
For Heter. Transformer, a \verb|class| token is appended to extract graph-level representations. The transformer is optimized using the match outcome loss $\mathcal{L}_{out}$. The Heter. GCN is pre-trained separately by applying average pooling over processed graph embeddings to obtain graph-level representations, followed by the same match outcome loss. By pre-training on individual matches, the Player Interaction Network receives stronger supervision, enabling it to attribute match outcomes to individual players more effectively.

In Stage 2, we train the remaining model components, which focus on fusing global and local player information, capturing team interactions and aggregating historical match information. To enhance efficiency, we freeze the Player Interaction Network (except for the gating network) and use pre-trained player embeddings from past matches, which can be computed offline for faster training. The match outcome loss $\mathcal{L}_{\text{out}}(y^i, \hat{y}^i)$ is then used to optimize the gating network, the Team Interaction Network and the Match Comparison Transformer. The Stage 2 ensures that the model learns team-level dynamics and historical match trends in an efficient and structured manner.

\section{Experiments}

\subsection{Experimental Settings} \label{setup}
\noindent\textbf{Dataset.}
To evaluate our method, we train and test on the WyScout Open Access Dataset \cite{2019wyscout} with detailed statistics in Table \ref{tab:dataset}.
It encompasses data from 1941 matches spanning 7 top-tier competitions, including the 2017/18 seasons of the Spanish(SPA), English (ENG), Italian (ITA), German (GER), and French (FRE) men's first divisions, along with additional data from the 2016 European Cup (EC) and the 2018 World Cup (WC). This dataset offers comprehensive event annotations, enabling the analysis and exploitation of potential interactive relationships during the match. 
Specifically, the dataset is split into 80\% training data and 20\% test samples for each division, sorted by time. 
The label distributions [win, draw, lose] are $[695, 397, 460]$ (train) and $[177, 80, 132]$ (test) respectively. Weighted random sampler is adapted to alleviate the impact of imbalance labels. We pre-processed the dataset into a match-level dataset containing date, match ID, outcome, team ID, coach ID, and player IDs. Additionally, we construct an interaction graph per match beforehand and maintain a historical match ID list for each player. During data loading, the model retrieves all necessary information including match-level features, historical context, and graph structure based on players' recent matches.

\begin{table}[t]
    \centering
    \caption{Statistics of WyScout Open Access Dataset ~\cite{2019wyscout}.}
    \resizebox{\columnwidth}{!}{
    \begin{tabular}{p{2cm}| c c c c}
    \hline Competition & Matches & Players & Teams & Events \\
    \hline SPA & 380 & 517 & 20 & 628,659 \\
    \hline ENG & 380 & 477 & 20 & 643,150 \\
    \hline ITA & 380 & 492 & 20 & 647,372 \\
    \hline GER & 306 & 428 & 18 & 519,407 \\
    \hline FRE & 380 & 491 & 20 & 632,807 \\
    \hline EC & 51 & 377 & 24 & 78,140 \\
    \hline WC & 64 & 511 & 32 & 101,759 \\
    \hline \textbf{Total} & $\mathbf{1,941}$ & $\mathbf{3293}$ & $\mathbf{154}$ & $\mathbf{3,251,294}$ \\
    \hline
    \end{tabular}
    }
    \label{tab:dataset}
\end{table}

\noindent\textbf{Baselines.} To evaluate the performance of match outcome prediction, we compare our model with the following baselines: 
\begin{itemize}[leftmargin=*]
    \item \textbf{MLP} ~\cite{soccer2020mlp} is a soccer match outcome prediction framework using fully connected neural networks with hand-crafted player/team related features as input. We adapt a similar MLP-based prediction network but with historical event counts and player attributes (height, weight, age, role) as player features like our proposed method for a fair comparison. 
    
    \item \textbf{RNN} ~\cite{soccer2018rnn} is adapting a many-to-one recurrent neural network to predict the next soccer match's outcome with the information of past matches. Similarly, we use the same features as those of MLP to construct the input instead and formulate a sequence of past matches for the recurrent structure. 
    
    \item \textbf{P-Graph} ~\cite{anzer2022playergraph} is a graph-based method that represents multi-player trajectories as a graph, with players as nodes, to predict tactical patterns in soccer. We adapt this approach for outcome prediction by constructing the player graph as described in Section\ref{sec:player} and applying the same Graph Attention Network (GAT) with an added classification head to predict match outcomes.

    \item \textbf{T-Graph} ~\cite{zhao2023teamgraph} is a graph-based approach that models win/loss relationships in basketball as a graph with teams as nodes for outcome prediction. We adapt this idea to soccer by constructing team graphs as described in Section\ref{sec:team} and applying a Graph Attention Network (GAT) with a classification head to predict match outcomes.
    
    \item \textbf{DraftRec} ~\cite{2022draftrec} is a transformer-based method for draft recommendation and match prediction in online games. It models player histories as temporal sequences and aggregates them into team representations using a transformer architecture. We adapt it for soccer outcome prediction by modifying the input features and removing the draft component. In contrast to DraftRec, our method explicitly models both historical player–player and team–team interactions as graphs to better capture the structural dynamics of soccer matches.
\end{itemize}

\noindent\textbf{Evaluation Metrics.} To quantitatively assess the performance of the proposed method, we utilize accuracy, a standard evaluation metric for multi-class classification. 
The accuracy is further separated into \textit{win}, \textit{draw}, \textit{lose} and \textit{total} for a more comprehensive analysis of performance.  

\noindent\textbf{Implementation Details.}
In regard to network architecture, Heter. Transformer, Heter. GCN, and the GCN of Team Interaction Network all contain $3$ layers, with the hidden dimension initialized to $64$ and the output embedding set to $16$. 
On top of them, the Match Comparison Transformer has $1$ transformer encoder layer.
In terms of match history length $T$, we experimented ${5, 10, 20}$ and empirically chose $10$ to train our final model. 
Adam ~\cite{kingma2014adam} optimizer is used with learning rate set to $1e-3$ for model optimization. The optimization is conducted for 2328 steps in Stage 1 and 2134 steps in Stage 2. 
When determining the outcome, we empirically set the output value range for each class: $[0, 4/7]$ as a home lose, $[4/7, 5/7]$ as a draw, and $[5/7, 1]$ as a home win due to difficulty in draw outcome prediction.
We employ the identical feature set for both the MLP and the RNN models. In the case of the MLP, we concatenate all historical features and feed them collectively into the network. Conversely, with the RNN, we input the historical match information sequentially to forecast the subsequent outcome.

\subsection{Results}

\begin{table*}[t]
    \centering
    \caption{Quantitative evaluation on the WyScout Open Access Dataset ~\cite{2019wyscout}. The values presented are accuracy (\%) and the higher is the better. Bold indicates the best performance and underline refers to the second best one.}
    
    \begin{subtable}[c]{\textwidth}
    \setlength{\tabcolsep}{4pt}
    \begin{tabular*}{\textwidth}{@{\extracolsep{\fill}} 
    l 
    p{0.6cm}<{\centering}p{0.6cm}<{\centering}p{0.6cm}<{\centering}p{0.6cm}<{\centering} 
    p{0.6cm}<{\centering}p{0.6cm}<{\centering}p{0.6cm}<{\centering}p{0.6cm}<{\centering} 
    p{0.6cm}<{\centering}p{0.6cm}<{\centering}p{0.6cm}<{\centering}p{0.6cm}<{\centering} 
    p{0.6cm}<{\centering}p{0.6cm}<{\centering}p{0.6cm}<{\centering}p{0.8cm}<{\centering} 
    @{}}
    \hline \multirow{2}{*}{ Methods } & \multicolumn{4}{c}{SPA} &  \multicolumn{4}{c}{ENG} & \multicolumn{4}{c}{ITA} & \multicolumn{4}{c}{GER} \\
    \cline { 2 - 5 } \cline {6 - 9} \cline {10 - 13} \cline {14 - 17} 
    & Win & Draw & Lose & Avg & Win & Draw & Lose & Avg & Win & Draw & Lose & Avg & Win & Draw & Lose & Avg \\
    \hline
    MLP ~\cite{soccer2020mlp} & 48.28 & 37.50 & 22.22 & 38.10 & 51.72 & 35.00 & 60.00 & 50.00 & 52.78 & 58.33 & 46.15 & 52.33 & 54.84 & 35.00 & 42.11 & 45.71\\
    
    RNN ~\cite{soccer2018rnn} & 62.07 & 43.75 & 33.33 & \textbf{49.21} & 44.83 & 20.00 & 64.00 & 44.59 & 61.11 & 41.67 & 38.46 & 48.84 & 54.84 & 40.00 & 52.63 & \textbf{50.00}\\

    P-Graph ~\cite{anzer2022playergraph} & 57.69 & 12.50 & 50.00 & 44.74 & 64.00 & 26.32 & 50.00 & 48.68 & 62.50 & 29.41 & 65.71 & \underline{56.58} & 40.00 & 23.08 & 38.24 & 35.48 \\

    T-Graph ~\cite{zhao2023teamgraph} & 53.85 & 6.25 & 55.88 & 44.74 & 56.00 & 10.53 & 62.50 & 48.68 & 62.50 & 17.65 & 62.86 & 52.63 & 53.33 & 0.00 & 50.00 & 40.32 \\
    
    DraftRec ~\cite{2022draftrec} & 62.07 & 18.75 & 50.00 & \underline{47.62} & 62.07 & 30.00 & 60.00 & \underline{52.70} & 61.11 & 33.33 & 65.39 & 54.65 & 45.16 & 20.00 & 63.16 & 42.86\\
    \hline
    
    Ours & 51.72 & 31.25 & 61.11 & \textbf{49.21} & 65.52 & 30.00 & 72.00 & \textbf{58.11} & 69.44 & 16.67 & 76.92& \textbf{56.98}& 48.39 & 30.00 & 68.42 & \underline{48.57} \\
    
    \hline
    \end{tabular*}
    \end{subtable}
    
    \vspace{10px}
    
    \begin{subtable}[c]{\textwidth}
    \setlength{\tabcolsep}{4pt}
    \begin{tabular*}{\textwidth}{@{\extracolsep{\fill}} l cccc cccc cccc @{}}
    \hline \multirow{2}{*}{ Methods } & \multicolumn{4}{c}{FRE} & \multicolumn{4}{c}{EC+WC} & \multicolumn{4}{c}{Total} \\
    \cline { 2 - 5 } \cline {6 - 9} \cline {10 - 13}
    & Win & Draw & Lose & Avg & Win & Draw & Lose & Avg & Win & Draw & Lose & Avg \\
    \hline
    MLP ~\cite{soccer2020mlp} & 57.69 & 30.00 & 57.69 & 50.00 & 66.67 & 0.00 & 33.33 & 33.33 & 53.50 & \textbf{37.74} & 46.03 & 46.79 \\
    RNN ~\cite{soccer2018rnn} & 65.38 & 30.00 & 53.85 & \underline{51.39} & 66.67 & 0.00 & 16.67 & 27.27 & \underline{57.96} & \underline{33.02} & 46.03 & 47.30 \\
    P-Graph ~\cite{anzer2022playergraph} & 68.18 & 35.00 & 52.94 & \textbf{52.63} & 33.33 & 28.57 & 54.55 & \underline{41.67} & \textbf{58.47} & 26.09 & 51.67 & 47.69\\
    T-Graph ~\cite{zhao2023teamgraph} & 33.33 & 0.00 & 60.00 & 36.36 & 33.33 & 0.00 & 54.55 & 33.33 & \textbf{58.47} & 9.78 & \underline{58.33} & 46.92 \\
    DraftRec ~\cite{2022draftrec} & 61.54 & 15.00 & 61.54 & 48.61 & 33.33 & 33.33 & 25.00 & 29.17 & 57.33 & 24.53 & 57.14 & \underline{48.33}\\
    \hline
    Ours & 53.85 & 15.00 & 69.23 & 48.61 & 50.00 & 25.00 & 66.67 & \textbf{45.83} & \underline{57.96} & 24.53 & \textbf{68.25} & \textbf{52.19} \\
    \hline
    \end{tabular*}
    \end{subtable}
    \label{tab:result1}
\end{table*}

Table~\ref{tab:result1} summarizes the quantitative evaluation results. To ensure stability, we combine the results for EC and WC due to the limited number of test samples. Overall, our method consistently outperforms all baselines in terms of average accuracy, achieving $52.19\%$. It achieves the second-best performance in win prediction at $57.96\%$, which is comparable to the top result, and the highest performance in lose prediction at $68.25\%$. These results demonstrate the effectiveness of our approach in capturing match characteristics by jointly modeling player and team interactions. Across divisions, our method outperforms the baselines in most cases, and still performs competitively in GER and FRE.

However, while HIGFormer performs well in predicting win and lose, it does not achieve the same level of accuracy in draw prediction compared to the baseline models. It is noticed that even the highest accuracy achieved for draw prediction among all methods is only $37\%$, significantly lower than the accuracy for the other two classes. The intricacy associated with forecasting match draws is not confined to our model but rather represents a standing challenge within the domain of soccer match prediction, as acknowledged in previous literature ~\cite{soccer2015ma,soccer2018rnn,2019boost,beal2021baseline}. The challenge may arise from accurately approximating the distribution of draw outcomes, as their characteristics can be somewhat inherently ambiguous, lying between those of wins and losses.  

\subsection{Ablation Study}

\begin{table}[t]
    \centering
    \caption{Quantitative results of ablation study on network components, loss settings and training strategy.}
    \begin{tabular}{p{3.3cm} cccc}
    \hline \multirow{2}{*}{ Methods } & \multicolumn{4}{c}{Total} \\
    \cline { 2 - 5 }
    & Win & Draw & Lose & Avg \\
    \hline
    Ours & \textbf{57.96} & 24.53 & \textbf{68.25} & \textbf{52.19} \\
    \hline
    w/o Heter. GCN & 54.14 & 28.30 & 63.49 & 50.13 \\
    w/o Heter. Transformer & \underline{55.41} & 24.53 & 61.11 & 48.84 \\
    w/o Player Inter. Net. & 52.87 & \textbf{32.08} & 52.87 & 48.59 \\
    w/o Team Inter. Net. & 53.50 & \underline{30.19} & 58.73 & 48.84 \\
    \hline
    w $[0,2/5,3/5,1]$ & 55.30 & 27.50 & 57.06 & 50.39 \\
    w $[0,1/3,2/3,1]$ & 52.27 & 32.50 & 57.06 & 48.84 \\
    w cross-entropy & 53.03 & 17.50 & \underline{64.97} & \underline{51.16} \\
    \hline
    w/o two-stage training & 54.24 & 21.74 & 57.22 & 47.95 \\
    \hline
    \end{tabular}
    \label{tab:ablat}
\end{table}

To validate the proposed mechanisms of HIGFormer, the ablation study was conducted by removing four proposed components from the complete pipeline, namely heterogeneous graph convolution network, heterogeneous graph transformer, Player Interaction Network and Team Interaction Network. The corresponding quantitative evaluation results are shown in Table \ref{tab:ablat}. 

\noindent\textbf{Heterogeneous Graph Convolution Network (Heter. GCN)} was removed from the Player Interaction Network. Specifically, the player embedding is only obtained by the heterogeneous graph transformer as in Equation (\ref{eqa:global}). After removing the local branch from Player Interaction Network, prediction accuracy has dropped across all the outcomes except draw, which reflects the value of incorporating local graph information learning on player interaction. 

\noindent\textbf{Heterogeneous Graph Transformer (Heter. Transformer)} was altered from the Player Interaction Network. Particularly, only Equation (\ref{eqa:local}) is used to obtain player embedding. In pursuit of this removal, the Player Interaction Network is dedicated to address the local graph learning only. The results show that the model suffers from even greater performance drop, indicating the unique value of learning player interactions across graphs globally. 

\noindent\textbf{Player Interaction Network (Player Inter. Net.)} was removed to investigate the importance of player interaction information learning. The Match Comparison Transformer now utilizes solely the team interaction information, leaving the exploration of interactions among players in the past matches unaddressed. Interestingly, despite a significant decrease in performance compared to our final model, the predictive capability remains relatively acceptable. This suggests that even simple team history information could capture certain insights into future match outcomes. 

\noindent\textbf{Team Interaction Network (Team Inter. Net.)} was removed from the pipeline, which results in the Match Comparison Transformer accepting only player embeddings as input and representing team-level comparisons solely based on player information. As anticipated, the corresponding performance deteriorates, confirming our expectation that soccer teams are enduring entities with distinct advantages. This stands in contrast to the scenarios where teams are formed dynamically, such as in online games. 

\noindent\textbf{Different Loss Setting.} We provide additional analysis on different loss setting in addition to the MSE with $[1,4/7,5/7,1]$ configuration. Results indicate that cross-entropy loss yields poor draw prediction accuracy, likely due to the inherent difficulty of modeling draws. In contrast, our empirical MSE formulation achieves the best overall performance by capturing the ordinal relationships among outcomes. 

\noindent\textbf{Two-Stage Training.} We evaluate the effectiveness of the two-stage training strategy. In the one-stage setting, each match contains up to 460 graphs (based on 10 historical matches per player), which complicates graph learning. As a result, the model performs similarly to DraftRec, relying more on non-graph features and resembling a transformer-only architecture. 

\subsection{Attention Weights Analysis}

\begin{figure}[t]
    \centering
    \includegraphics[width=\linewidth]{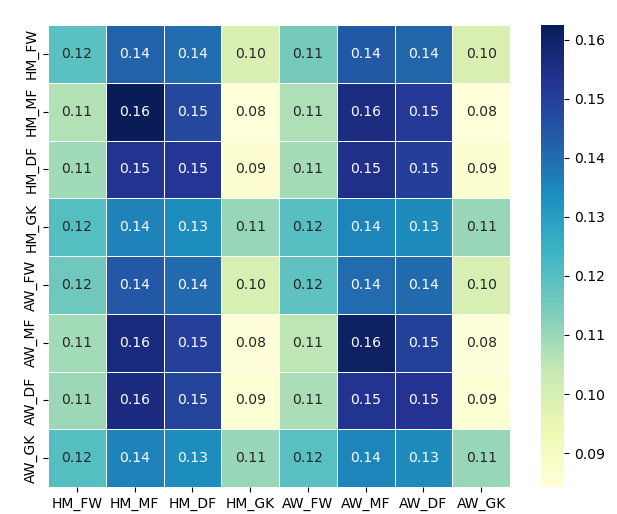}
    \caption{Visualization of the attention weights for Player Interaction Network, which are ordered by roles of players.}
    \label{fig:player-att}
\end{figure}

To better understand how HIGFormer learns interaction relationships between different player roles, we visualize the attention weights in the final self-attention layer of the heterogeneous transformer in the Player Interaction Network. Specifically, we compute the average attention weights from the test data in the Wyscout Open Access Dataset \cite{2019wyscout} and visualize them in Figure \ref{fig:player-att}. We remove attention associated with edge tokens and include only player-to-player attention with normalization for better comparison. The attention values are grouped by roles: forward (FW), midfielder (MF), and defender (DF) for both home (HM) and away (AW) teams.

Figure \ref{fig:player-att} reveals key patterns in player interactions. 
First, players tend to assign higher attention to others in the same role, particularly midfielders (MF) and defenders (DF), which is logical given their proximity on the field.
Second, midfielders and defenders receive the most attention, both within and across teams, suggesting their crucial role in game control and defense, which significantly impacts match outcomes.
Finally, goalkeepers (GK) consistently receive lower attention scores, likely due to their limited direct interactions with other players, as they engage in fewer events compared to outfield players.
These findings indicate that HIGFormer effectively captures meaningful interaction patterns, highlighting the strategic importance of midfield and defensive structures while correctly modeling the reduced involvement of goalkeepers in player interactions.

\subsection{Application to Player Evaluation}

\begin{table}[t]
    \centering
    \caption{Outcome differences after player substitution for teams at different ranks}
    \begin{tabular}{p{4cm} ccc}
    \hline 
    Team $\rightarrow$ Player & Win & Draw & Lose \\
    \hline
    \textit{Real Betis Balompie} (top) & 25.00 & 59.38 & 15.63 \\
    $\rightarrow$ L. Messi & +7.14 & -2.24 & -4.92  \\
    $\rightarrow$ C. Ronaldo & +6.03 & -4.21 & -1.84  \\
    $\rightarrow$ T. Kroos & +6.03 & -4.21 & -1.84  \\
    $\rightarrow$ G. Piqué & +6.03 & -4.21 & -1.84  \\
    \hline
    \textit{Girona} (middle) & 20.69 & 51.72 & 27.59 \\
    $\rightarrow$ L. Messi & +8.94 & +11.24 & -20.18  \\
    $\rightarrow$ C. Ronaldo & +8.94 & +11.24  & -20.18 \\
    $\rightarrow$ T. Kroos & +7.88 & +12.51 & -20.40  \\
    $\rightarrow$ G. Piqué & +8.94 & +11.24 & -20.18  \\
    \hline
    \textit{Celta Vigo} (low) & 17.86 & 71.43 & 10.71 \\
    $\rightarrow$ L. Messi & +1.49 & -16.59 & +15.10  \\
    $\rightarrow$ C. Ronaldo & +0.89 & -18.30 & +17.42  \\
    $\rightarrow$ T. Kroos & +0.89 & -18.30 & +17.42  \\
    $\rightarrow$ G. Piqué & +2.83 & -12.81 & +9.98  \\
    \hline
    \end{tabular}
    \label{tab:value}
\end{table}

Beyond match outcome prediction, our proposed method inherently encodes a player’s historical performance into a compact embedding based on interactive relationships. Since the final match outcome prediction flexibly depends on player embeddings, we can evaluate player performance by substituting different player embeddings within a team. This allows us to analyze how individual player changes impact team performance, potentially guiding tactical decisions and player selection.

Table \ref{tab:value} presents the match prediction results where we manually replace selected player embeddings within a team. We evaluate this impact on test set of three Spanish First Division teams from the 2017/18 season—Real Betis Balompié (top-ranked), Girona (mid-ranked), and Celta Vigo (low-ranked). 
The substitute players are selected from the best lineup nominees of the 2017/18 season, which includes L. Messi (forward), C. Ronaldo (forward), T. Kroos (midfielder), and G. Piqué (defender). The results show that incorporating top players significantly boosts the predicted winning percentage, often by more than $6\%$, while also reducing predicted losing probabilities, except in the low-ranked team.
Notably, in lower-ranked teams, substituting a single high-profile player does not drastically improve performance, likely due to soccer’s team-oriented nature. Unlike individual sports, a single substitution may not compensate for weaker overall team dynamics and could diverge from the team's typical player distribution.
This analysis demonstrates the model’s ability to effectively assess player impact, offering valuable insights for team management, tactical adjustments and player recruitment.

\section{Limitations and Future Work}
Despite its strong predictive performance and practical applications, HIGFormer has limitations that suggest directions for future improvement.
First, its accuracy in predicting draws is lower than wins and losses, likely due to the inherent ambiguity of draw outcomes. A potential improvement is score-based prediction, where outcomes are predicted as real-valued goals instead of discrete classes.
Second, our model currently focuses only on passing and defensive events in the interaction graph. Expanding it to include attacking actions such as shooting and dribbling, or even more detailed game events \cite{2022seq2event}, could enhance predictive depth.
Finally, we only use basic player event counts as input features to demonstrate HIGFormer’s effectiveness. However, soccer outcome prediction research \cite{2022outcomereview} has introduced many domain-specific engineered features. Incorporating these could further boost prediction accuracy.

\section{Conclusion}
In this paper, we propose HIGFormer, a novel Player-Team Heterogeneous Interaction Graph Transformer for pre-match soccer outcome prediction which uniquely integrates multi-level player and team interactions.
HIGFormer devises a Player Interaction Network to model the historical player information as heterogeneous interaction graph and encode player performance with the mixture of global graph transformer and local graph convolution. 
Additional, Team Interaction Network encodes historical team performance by exploiting a winning-rate based interaction graph. 
Finally, we introduce a Match Comparison Transformer to predict the match outcome based on integrated player embeddings and team embeddings. 
Extensive experiments demonstrate the superior predictive performance of HIGFormer without heavy feature engineering and reveals useful applications in player performance evaluation.

\begin{acks}
This study was partially supported by Australian Research Council (ARC) grant \#LP230100294 and \#DP240101353.
\end{acks}


\bibliographystyle{ACM-Reference-Format}
\balance
\input{mybib.bbl}

\clearpage

\appendix

\section{Symbol Reference Table}

To aid understanding of our proposed method, we provide a symbol reference table (Table~\ref{tab:symbols}) that clarifies the mathematical notations used throughout the paper.

\begin{table}[h]
  \centering
  \caption{Symbol Reference Table}
  \label{tab:symbols}
  \begin{tabular}{@{}cl@{}}
    \toprule
    \textbf{Symbol} & \textbf{Description} \\
    \midrule
    $i$         & Match index \\
    $y$         & Match outcome \\
    $\hat{y}$   & Predicted match outcome \\
    $R$         & Team red \\
    $B$         & Team blue \\
    $p$         & Match player \\
    $n$         & Player index \\
    $\mathbf{h}$      & Historical information of a past match \\
    $\mathbf{H}$  & Historical information of past matches \\
    $\mathbf{c}$         & Key event counts of a match \\
    $f_{\theta}$         & A learnable function with parameters $\theta$ \\
    $G$         & An interaction graph \\
    $V$         & All nodes of an interaction graph \\
    $v$         & A node of an interaction graph \\
    $E$         & All edges of an interaction graph \\
    $e$         & An edge of an interaction graph \\
    $A$         & Possible node types \\
    $K$         & Possible edge types \\
    $k$         & An edge type \\
    $\mathbf{X}$         & Features matrix \\
    $\mathbf{x}$         & Features vector \\
    $\mathbb{R}$     & Set of real numbers \\
    $d$         & Latent feature dimension \\
    $\mathbf{P}$         & A learnable embedding \\
    $Enc$         & Encoder network \\
    $\mathbf{Z}$         & Latent features matrix \\
    $\mathbf{z}$         & Latent features vector \\
    $\mathcal{N}(\cdot)$         & Local neighborhood of a node \\
    $\alpha$         &  Attention score\\
    $\mathbf{a}$         & Learnable weight vector \\
    $\mathbf{W}$         & Learnable weight matrix \\
    $\mathbf{p}$         & Expert weight vector \\
    $T$         & Number of past matches selected \\
    $t$         & Selected past match index \\
    $\mathbf{r}$         & Red team embedding vector \\
    $\mathbf{b}$         & Blue team embedding vector \\
    \bottomrule
  \end{tabular}
\end{table} 

\end{document}

%% file: mybib.bbl

%% file: soccer.bbl
\begin{thebibliography}{66}


\ifx \showCODEN    \undefined \def \showCODEN     #1{\unskip}     \fi
\ifx \showISBNx    \undefined \def \showISBNx     #1{\unskip}     \fi
\ifx \showISBNxiii \undefined \def \showISBNxiii  #1{\unskip}     \fi
\ifx \showISSN     \undefined \def \showISSN      #1{\unskip}     \fi
\ifx \showLCCN     \undefined \def \showLCCN      #1{\unskip}     \fi
\ifx \shownote     \undefined \def \shownote      #1{#1}          \fi
\ifx \showarticletitle \undefined \def \showarticletitle #1{#1}   \fi
\ifx \showURL      \undefined \def \showURL       {\relax}        \fi
\providecommand\bibfield[2]{#2}
\providecommand\bibinfo[2]{#2}
\providecommand\natexlab[1]{#1}
\providecommand\showeprint[2][]{arXiv:#2}

\bibitem[mar(2023)]%
        {market}
 \bibinfo{year}{2023}\natexlab{}.
\newblock \bibinfo{title}{Annual Review of Football Finance}.
\newblock \bibinfo{howpublished}{\url{https://www2.deloitte.com/uk/en/pages/sports-business-group/articles/annual-review-of-football-finance-europe.html}}.
\newblock
\newblock
\shownote{Accessed: 2024-04-29}.


\bibitem[AlMulla et~al\mbox{.}(2023)]%
        {soccer2023rnn}
\bibfield{author}{\bibinfo{person}{Jassim AlMulla}, \bibinfo{person}{Mohammad~Tariqul Islam}, \bibinfo{person}{Hamada~RH Al-Absi}, {and} \bibinfo{person}{Tanvir Alam}.} \bibinfo{year}{2023}\natexlab{}.
\newblock \showarticletitle{SoccerNet: A Gated Recurrent Unit-based model to predict soccer match winners}.
\newblock \bibinfo{journal}{\emph{Plos One}} \bibinfo{volume}{18}, \bibinfo{number}{8} (\bibinfo{year}{2023}), \bibinfo{pages}{e0288933}.
\newblock


\bibitem[Alon and Yahav(2020)]%
        {2020oversquashing}
\bibfield{author}{\bibinfo{person}{Uri Alon} {and} \bibinfo{person}{Eran Yahav}.} \bibinfo{year}{2020}\natexlab{}.
\newblock \showarticletitle{On the bottleneck of graph neural networks and its practical implications}.
\newblock \bibinfo{journal}{\emph{arXiv preprint arXiv:2006.05205}} (\bibinfo{year}{2020}).
\newblock


\bibitem[Anzer et~al\mbox{.}(2022)]%
        {anzer2022playergraph}
\bibfield{author}{\bibinfo{person}{Gabriel Anzer}, \bibinfo{person}{Pascal Bauer}, \bibinfo{person}{Ulf Brefeld}, {and} \bibinfo{person}{Dennis Fa{\ss}meyer}.} \bibinfo{year}{2022}\natexlab{}.
\newblock \showarticletitle{Detection of tactical patterns using semi-supervised graph neural networks}. In \bibinfo{booktitle}{\emph{16th MIT Sloan Sports Analytics Conference}}. \bibinfo{pages}{1--15}.
\newblock


\bibitem[Arabzad et~al\mbox{.}(2014)]%
        {soccer2014mlp}
\bibfield{author}{\bibinfo{person}{S~Mohammad Arabzad}, \bibinfo{person}{Mohamad~Ebrahim Tayebi~Araghi}, \bibinfo{person}{Soheil Sadi-Nezhad}, {and} \bibinfo{person}{Nooshin Ghofrani}.} \bibinfo{year}{2014}\natexlab{}.
\newblock \showarticletitle{Football match results prediction using artificial neural networks; the case of Iran Pro League}.
\newblock \bibinfo{journal}{\emph{Journal of Applied Research on Industrial Engineering}} \bibinfo{volume}{1}, \bibinfo{number}{3} (\bibinfo{year}{2014}), \bibinfo{pages}{159--179}.
\newblock


\bibitem[Beal et~al\mbox{.}(2021)]%
        {beal2021baseline}
\bibfield{author}{\bibinfo{person}{Ryan Beal}, \bibinfo{person}{Stuart~E Middleton}, \bibinfo{person}{Timothy~J Norman}, {and} \bibinfo{person}{Sarvapali~D Ramchurn}.} \bibinfo{year}{2021}\natexlab{}.
\newblock \showarticletitle{Combining machine learning and human experts to predict match outcomes in football: A baseline model}. In \bibinfo{booktitle}{\emph{Proceedings of the AAAI Conference on Artificial Intelligence}}, Vol.~\bibinfo{volume}{35}. \bibinfo{pages}{15447--15451}.
\newblock


\bibitem[Berrar et~al\mbox{.}(2019)]%
        {2019domain}
\bibfield{author}{\bibinfo{person}{Daniel Berrar}, \bibinfo{person}{Philippe Lopes}, {and} \bibinfo{person}{Werner Dubitzky}.} \bibinfo{year}{2019}\natexlab{}.
\newblock \showarticletitle{Incorporating domain knowledge in machine learning for soccer outcome prediction}.
\newblock \bibinfo{journal}{\emph{Machine Learning}}  \bibinfo{volume}{108} (\bibinfo{year}{2019}), \bibinfo{pages}{97--126}.
\newblock


\bibitem[Bunker and Susnjak(2022)]%
        {2022outcomereview}
\bibfield{author}{\bibinfo{person}{Rory Bunker} {and} \bibinfo{person}{Teo Susnjak}.} \bibinfo{year}{2022}\natexlab{}.
\newblock \showarticletitle{The application of machine learning techniques for predicting match results in team sport: A review}.
\newblock \bibinfo{journal}{\emph{Journal of Artificial Intelligence Research}}  \bibinfo{volume}{73} (\bibinfo{year}{2022}), \bibinfo{pages}{1285--1322}.
\newblock


\bibitem[Chen et~al\mbox{.}(2022)]%
        {2022subgraphsat}
\bibfield{author}{\bibinfo{person}{Dexiong Chen}, \bibinfo{person}{Leslie O’Bray}, {and} \bibinfo{person}{Karsten Borgwardt}.} \bibinfo{year}{2022}\natexlab{}.
\newblock \showarticletitle{Structure-aware transformer for graph representation learning}. In \bibinfo{booktitle}{\emph{International Conference on Machine Learning}}. PMLR, \bibinfo{pages}{3469--3489}.
\newblock


\bibitem[Constantinou(2019)]%
        {2019bayesian}
\bibfield{author}{\bibinfo{person}{Anthony~C Constantinou}.} \bibinfo{year}{2019}\natexlab{}.
\newblock \showarticletitle{Dolores: a model that predicts football match outcomes from all over the world}.
\newblock \bibinfo{journal}{\emph{Machine Learning}} \bibinfo{volume}{108}, \bibinfo{number}{1} (\bibinfo{year}{2019}), \bibinfo{pages}{49--75}.
\newblock


\bibitem[Constantinou and Fenton(2013)]%
        {2013pirating}
\bibfield{author}{\bibinfo{person}{Anthony~Costa Constantinou} {and} \bibinfo{person}{Norman~Elliott Fenton}.} \bibinfo{year}{2013}\natexlab{}.
\newblock \showarticletitle{Determining the level of ability of football teams by dynamic ratings based on the relative discrepancies in scores between adversaries}.
\newblock \bibinfo{journal}{\emph{Journal of Quantitative Analysis in Sports}} \bibinfo{volume}{9}, \bibinfo{number}{1} (\bibinfo{year}{2013}), \bibinfo{pages}{37--50}.
\newblock


\bibitem[Danisik et~al\mbox{.}(2018)]%
        {soccer2018rnn}
\bibfield{author}{\bibinfo{person}{Norbert Danisik}, \bibinfo{person}{Peter Lacko}, {and} \bibinfo{person}{Michal Farkas}.} \bibinfo{year}{2018}\natexlab{}.
\newblock \showarticletitle{Football match prediction using players attributes}. In \bibinfo{booktitle}{\emph{2018 World Symposium on Digital Intelligence for Systems and Machines}}. IEEE, \bibinfo{pages}{201--206}.
\newblock


\bibitem[Dubitzky et~al\mbox{.}(2019)]%
        {2019challenge}
\bibfield{author}{\bibinfo{person}{Werner Dubitzky}, \bibinfo{person}{Philippe Lopes}, \bibinfo{person}{Jesse Davis}, {and} \bibinfo{person}{Daniel Berrar}.} \bibinfo{year}{2019}\natexlab{}.
\newblock \showarticletitle{The open international soccer database for machine learning}.
\newblock \bibinfo{journal}{\emph{Machine Learning}}  \bibinfo{volume}{108} (\bibinfo{year}{2019}), \bibinfo{pages}{9--28}.
\newblock


\bibitem[Dwivedi and Bresson(2020)]%
        {2020gt}
\bibfield{author}{\bibinfo{person}{Vijay~Prakash Dwivedi} {and} \bibinfo{person}{Xavier Bresson}.} \bibinfo{year}{2020}\natexlab{}.
\newblock \showarticletitle{A generalization of transformer networks to graphs}.
\newblock \bibinfo{journal}{\emph{arXiv preprint arXiv:2012.09699}} (\bibinfo{year}{2020}).
\newblock


\bibitem[Fedus et~al\mbox{.}(2022)]%
        {2022switchtransformer}
\bibfield{author}{\bibinfo{person}{William Fedus}, \bibinfo{person}{Barret Zoph}, {and} \bibinfo{person}{Noam Shazeer}.} \bibinfo{year}{2022}\natexlab{}.
\newblock \showarticletitle{Switch transformers: Scaling to trillion parameter models with simple and efficient sparsity}.
\newblock \bibinfo{journal}{\emph{Journal of Machine Learning Research}} \bibinfo{volume}{23}, \bibinfo{number}{120} (\bibinfo{year}{2022}), \bibinfo{pages}{1--39}.
\newblock


\bibitem[Fu et~al\mbox{.}(2020)]%
        {2020magnn}
\bibfield{author}{\bibinfo{person}{Xinyu Fu}, \bibinfo{person}{Jiani Zhang}, \bibinfo{person}{Ziqiao Meng}, {and} \bibinfo{person}{Irwin King}.} \bibinfo{year}{2020}\natexlab{}.
\newblock \showarticletitle{Magnn: Metapath aggregated graph neural network for heterogeneous graph embedding}. In \bibinfo{booktitle}{\emph{Proceedings of the Web Conference 2020}}. \bibinfo{pages}{2331--2341}.
\newblock


\bibitem[Gong et~al\mbox{.}(2020)]%
        {2020optmatch}
\bibfield{author}{\bibinfo{person}{Linxia Gong}, \bibinfo{person}{Xiaochuan Feng}, \bibinfo{person}{Dezhi Ye}, \bibinfo{person}{Hao Li}, \bibinfo{person}{Runze Wu}, \bibinfo{person}{Jianrong Tao}, \bibinfo{person}{Changjie Fan}, {and} \bibinfo{person}{Peng Cui}.} \bibinfo{year}{2020}\natexlab{}.
\newblock \showarticletitle{Optmatch: Optimized matchmaking via modeling the high-order interactions on the arena}. In \bibinfo{booktitle}{\emph{Proceedings of the 26th ACM SIGKDD International Conference on Knowledge Discovery \& Data Mining}}. \bibinfo{pages}{2300--2310}.
\newblock


\bibitem[Gu et~al\mbox{.}(2021)]%
        {2021neuralac}
\bibfield{author}{\bibinfo{person}{Yin Gu}, \bibinfo{person}{Qi Liu}, \bibinfo{person}{Kai Zhang}, \bibinfo{person}{Zhenya Huang}, \bibinfo{person}{Runze Wu}, {and} \bibinfo{person}{Jianrong Tao}.} \bibinfo{year}{2021}\natexlab{}.
\newblock \showarticletitle{Neuralac: Learning cooperation and competition effects for match outcome prediction}. In \bibinfo{booktitle}{\emph{Proceedings of the AAAI Conference on Artificial Intelligence}}, Vol.~\bibinfo{volume}{35}. \bibinfo{pages}{4072--4080}.
\newblock


\bibitem[Gu et~al\mbox{.}(2023)]%
        {2023massne}
\bibfield{author}{\bibinfo{person}{Yin Gu}, \bibinfo{person}{Kai Zhang}, \bibinfo{person}{Qi Liu}, \bibinfo{person}{Xin Lin}, \bibinfo{person}{Zhenya Huang}, {and} \bibinfo{person}{Enhong Chen}.} \bibinfo{year}{2023}\natexlab{}.
\newblock \showarticletitle{MassNE: Exploring Higher-Order Interactions with Marginal Effect for Massive Battle Outcome Prediction}. In \bibinfo{booktitle}{\emph{Proceedings of the ACM Web Conference 2023}}. \bibinfo{pages}{2710--2718}.
\newblock


\bibitem[Guan and Wang(2022)]%
        {soccer2022mlp}
\bibfield{author}{\bibinfo{person}{Shuo Guan} {and} \bibinfo{person}{Xiaochen Wang}.} \bibinfo{year}{2022}\natexlab{}.
\newblock \showarticletitle{Optimization analysis of football match prediction model based on neural network}.
\newblock \bibinfo{journal}{\emph{Neural Computing and Applications}} (\bibinfo{year}{2022}), \bibinfo{pages}{1--17}.
\newblock


\bibitem[Hamilton et~al\mbox{.}(2017)]%
        {2017graphsage}
\bibfield{author}{\bibinfo{person}{Will Hamilton}, \bibinfo{person}{Zhitao Ying}, {and} \bibinfo{person}{Jure Leskovec}.} \bibinfo{year}{2017}\natexlab{}.
\newblock \showarticletitle{Inductive representation learning on large graphs}.
\newblock \bibinfo{journal}{\emph{Advances in Neural Information Processing Systems}}  \bibinfo{volume}{30} (\bibinfo{year}{2017}).
\newblock


\bibitem[Herbrich et~al\mbox{.}(2006)]%
        {2006trueskill}
\bibfield{author}{\bibinfo{person}{Ralf Herbrich}, \bibinfo{person}{Tom Minka}, {and} \bibinfo{person}{Thore Graepel}.} \bibinfo{year}{2006}\natexlab{}.
\newblock \showarticletitle{TrueSkill™: a Bayesian skill rating system}.
\newblock \bibinfo{journal}{\emph{Advances in Neural Information Processing Systems}}  \bibinfo{volume}{19} (\bibinfo{year}{2006}).
\newblock


\bibitem[Hodge et~al\mbox{.}(2019)]%
        {2019boostinggame}
\bibfield{author}{\bibinfo{person}{Victoria~J Hodge}, \bibinfo{person}{Sam Devlin}, \bibinfo{person}{Nick Sephton}, \bibinfo{person}{Florian Block}, \bibinfo{person}{Peter~I Cowling}, {and} \bibinfo{person}{Anders Drachen}.} \bibinfo{year}{2019}\natexlab{}.
\newblock \showarticletitle{Win prediction in multiplayer esports: Live professional match prediction}.
\newblock \bibinfo{journal}{\emph{IEEE Transactions on Games}} \bibinfo{volume}{13}, \bibinfo{number}{4} (\bibinfo{year}{2019}), \bibinfo{pages}{368--379}.
\newblock


\bibitem[Hu et~al\mbox{.}(2020)]%
        {2020hgt}
\bibfield{author}{\bibinfo{person}{Ziniu Hu}, \bibinfo{person}{Yuxiao Dong}, \bibinfo{person}{Kuansan Wang}, {and} \bibinfo{person}{Yizhou Sun}.} \bibinfo{year}{2020}\natexlab{}.
\newblock \showarticletitle{Heterogeneous graph transformer}. In \bibinfo{booktitle}{\emph{Proceedings of the Web Conference 2020}}. \bibinfo{pages}{2704--2710}.
\newblock


\bibitem[Huang and Chang(2010)]%
        {soccer2010mlp}
\bibfield{author}{\bibinfo{person}{Kou-Yuan Huang} {and} \bibinfo{person}{Wen-Lung Chang}.} \bibinfo{year}{2010}\natexlab{}.
\newblock \showarticletitle{A neural network method for prediction of 2006 world cup football game}. In \bibinfo{booktitle}{\emph{The 2010 International Joint Conference on Neural Networks}}. IEEE, \bibinfo{pages}{1--8}.
\newblock


\bibitem[Huang et~al\mbox{.}(2006)]%
        {2006bradley}
\bibfield{author}{\bibinfo{person}{Tzu-Kuo Huang}, \bibinfo{person}{Chih-Jen Lin}, {and} \bibinfo{person}{Ruby~C Weng}.} \bibinfo{year}{2006}\natexlab{}.
\newblock \showarticletitle{Ranking individuals by group comparisons}. In \bibinfo{booktitle}{\emph{Proceedings of the 23rd International Conference on Machine Learning}}. \bibinfo{pages}{425--432}.
\newblock


\bibitem[Hub{\'a}{\v{c}}ek et~al\mbox{.}(2019)]%
        {2019boost}
\bibfield{author}{\bibinfo{person}{Ond{\v{r}}ej Hub{\'a}{\v{c}}ek}, \bibinfo{person}{Gustav {\v{S}}ourek}, {and} \bibinfo{person}{Filip {\v{Z}}elezn{\`y}}.} \bibinfo{year}{2019}\natexlab{}.
\newblock \showarticletitle{Learning to predict soccer results from relational data with gradient boosted trees}.
\newblock \bibinfo{journal}{\emph{Machine Learning}}  \bibinfo{volume}{108} (\bibinfo{year}{2019}), \bibinfo{pages}{29--47}.
\newblock


\bibitem[Hucaljuk and Rakipovi{\'c}(2011)]%
        {soccer2011mlp}
\bibfield{author}{\bibinfo{person}{Josip Hucaljuk} {and} \bibinfo{person}{Alen Rakipovi{\'c}}.} \bibinfo{year}{2011}\natexlab{}.
\newblock \showarticletitle{Predicting football scores using machine learning techniques}. In \bibinfo{booktitle}{\emph{2011 Proceedings of the 34th International Convention MIPRO}}. IEEE, \bibinfo{pages}{1623--1627}.
\newblock


\bibitem[Jacobs et~al\mbox{.}(1991)]%
        {1991moe}
\bibfield{author}{\bibinfo{person}{Robert~A Jacobs}, \bibinfo{person}{Michael~I Jordan}, \bibinfo{person}{Steven~J Nowlan}, {and} \bibinfo{person}{Geoffrey~E Hinton}.} \bibinfo{year}{1991}\natexlab{}.
\newblock \showarticletitle{Adaptive mixtures of local experts}.
\newblock \bibinfo{journal}{\emph{Neural Computation}} \bibinfo{volume}{3}, \bibinfo{number}{1} (\bibinfo{year}{1991}), \bibinfo{pages}{79--87}.
\newblock


\bibitem[Joseph et~al\mbox{.}(2006)]%
        {soccer2006ma}
\bibfield{author}{\bibinfo{person}{Anito Joseph}, \bibinfo{person}{Norman~E Fenton}, {and} \bibinfo{person}{Martin Neil}.} \bibinfo{year}{2006}\natexlab{}.
\newblock \showarticletitle{Predicting football results using Bayesian nets and other machine learning techniques}.
\newblock \bibinfo{journal}{\emph{Knowledge-Based Systems}} \bibinfo{volume}{19}, \bibinfo{number}{7} (\bibinfo{year}{2006}), \bibinfo{pages}{544--553}.
\newblock


\bibitem[Kim et~al\mbox{.}(2022)]%
        {2022tokengt}
\bibfield{author}{\bibinfo{person}{Jinwoo Kim}, \bibinfo{person}{Dat Nguyen}, \bibinfo{person}{Seonwoo Min}, \bibinfo{person}{Sungjun Cho}, \bibinfo{person}{Moontae Lee}, \bibinfo{person}{Honglak Lee}, {and} \bibinfo{person}{Seunghoon Hong}.} \bibinfo{year}{2022}\natexlab{}.
\newblock \showarticletitle{Pure transformers are powerful graph learners}.
\newblock \bibinfo{journal}{\emph{Advances in Neural Information Processing Systems}}  \bibinfo{volume}{35} (\bibinfo{year}{2022}), \bibinfo{pages}{14582--14595}.
\newblock


\bibitem[Kingma and Ba(2014)]%
        {kingma2014adam}
\bibfield{author}{\bibinfo{person}{Diederik~P Kingma} {and} \bibinfo{person}{Jimmy Ba}.} \bibinfo{year}{2014}\natexlab{}.
\newblock \showarticletitle{Adam: A method for stochastic optimization}.
\newblock \bibinfo{journal}{\emph{arXiv preprint arXiv:1412.6980}} (\bibinfo{year}{2014}).
\newblock


\bibitem[Kipf and Welling(2016)]%
        {2016gcn}
\bibfield{author}{\bibinfo{person}{Thomas~N Kipf} {and} \bibinfo{person}{Max Welling}.} \bibinfo{year}{2016}\natexlab{}.
\newblock \showarticletitle{Semi-supervised classification with graph convolutional networks}.
\newblock \bibinfo{journal}{\emph{arXiv preprint arXiv:1609.02907}} (\bibinfo{year}{2016}).
\newblock


\bibitem[Kong et~al\mbox{.}(2023)]%
        {2023goat}
\bibfield{author}{\bibinfo{person}{Kezhi Kong}, \bibinfo{person}{Jiuhai Chen}, \bibinfo{person}{John Kirchenbauer}, \bibinfo{person}{Renkun Ni}, \bibinfo{person}{C~Bayan Bruss}, {and} \bibinfo{person}{Tom Goldstein}.} \bibinfo{year}{2023}\natexlab{}.
\newblock \showarticletitle{GOAT: A global transformer on large-scale graphs}. In \bibinfo{booktitle}{\emph{International Conference on Machine Learning}}. PMLR, \bibinfo{pages}{17375--17390}.
\newblock


\bibitem[Lazova and Basnarkov(2015)]%
        {2015pagerank}
\bibfield{author}{\bibinfo{person}{Verica Lazova} {and} \bibinfo{person}{Lasko Basnarkov}.} \bibinfo{year}{2015}\natexlab{}.
\newblock \showarticletitle{PageRank approach to ranking national football teams}.
\newblock \bibinfo{journal}{\emph{arXiv preprint arXiv:1503.01331}} (\bibinfo{year}{2015}).
\newblock


\bibitem[Lee et~al\mbox{.}(2022)]%
        {2022draftrec}
\bibfield{author}{\bibinfo{person}{Hojoon Lee}, \bibinfo{person}{Dongyoon Hwang}, \bibinfo{person}{Hyunseung Kim}, \bibinfo{person}{Byungkun Lee}, {and} \bibinfo{person}{Jaegul Choo}.} \bibinfo{year}{2022}\natexlab{}.
\newblock \showarticletitle{Draftrec: personalized draft recommendation for winning in multi-player online battle arena games}. In \bibinfo{booktitle}{\emph{Proceedings of the ACM Web Conference 2022}}. \bibinfo{pages}{3428--3439}.
\newblock


\bibitem[Li et~al\mbox{.}(2018)]%
        {2018hoi}
\bibfield{author}{\bibinfo{person}{Yao Li}, \bibinfo{person}{Minhao Cheng}, \bibinfo{person}{Kevin Fujii}, \bibinfo{person}{Fushing Hsieh}, {and} \bibinfo{person}{Cho-Jui Hsieh}.} \bibinfo{year}{2018}\natexlab{}.
\newblock \showarticletitle{Learning from group comparisons: exploiting higher order interactions}.
\newblock \bibinfo{journal}{\emph{Advances in Neural Information Processing Systems}}  \bibinfo{volume}{31} (\bibinfo{year}{2018}).
\newblock


\bibitem[Mao et~al\mbox{.}(2023)]%
        {2023hinormer}
\bibfield{author}{\bibinfo{person}{Qiheng Mao}, \bibinfo{person}{Zemin Liu}, \bibinfo{person}{Chenghao Liu}, {and} \bibinfo{person}{Jianling Sun}.} \bibinfo{year}{2023}\natexlab{}.
\newblock \showarticletitle{Hinormer: Representation learning on heterogeneous information networks with graph transformer}. In \bibinfo{booktitle}{\emph{Proceedings of the ACM Web Conference 2023}}. \bibinfo{pages}{599--610}.
\newblock


\bibitem[Martins et~al\mbox{.}(2017)]%
        {soccer2017mlp}
\bibfield{author}{\bibinfo{person}{Rodrigo~G Martins}, \bibinfo{person}{Alessandro~S Martins}, \bibinfo{person}{Leandro~A Neves}, \bibinfo{person}{Luciano~V Lima}, \bibinfo{person}{Edna~L Flores}, {and} \bibinfo{person}{Marcelo~Z do Nascimento}.} \bibinfo{year}{2017}\natexlab{}.
\newblock \showarticletitle{Exploring polynomial classifier to predict match results in football championships}.
\newblock \bibinfo{journal}{\emph{Expert Systems with Applications}}  \bibinfo{volume}{83} (\bibinfo{year}{2017}), \bibinfo{pages}{79--93}.
\newblock


\bibitem[Odachowski and Grekow(2013)]%
        {soccer2013ma}
\bibfield{author}{\bibinfo{person}{Karol Odachowski} {and} \bibinfo{person}{Jacek Grekow}.} \bibinfo{year}{2013}\natexlab{}.
\newblock \showarticletitle{Using bookmaker odds to predict the final result of football matches}. In \bibinfo{booktitle}{\emph{Knowledge Engineering, Machine Learning and Lattice Computing with Applications: 16th International Conference, KES 2012, San Sebastian, Spain, September 10-12, 2012, Revised Selected Papers 16}}. Springer, \bibinfo{pages}{196--205}.
\newblock


\bibitem[Oono and Suzuki(2019)]%
        {2019oversmoothing}
\bibfield{author}{\bibinfo{person}{Kenta Oono} {and} \bibinfo{person}{Taiji Suzuki}.} \bibinfo{year}{2019}\natexlab{}.
\newblock \showarticletitle{Graph neural networks exponentially lose expressive power for node classification}.
\newblock \bibinfo{journal}{\emph{arXiv preprint arXiv:1905.10947}} (\bibinfo{year}{2019}).
\newblock


\bibitem[Pappalardo et~al\mbox{.}(2019)]%
        {2019wyscout}
\bibfield{author}{\bibinfo{person}{Luca Pappalardo}, \bibinfo{person}{Paolo Cintia}, \bibinfo{person}{Alessio Rossi}, \bibinfo{person}{Emanuele Massucco}, \bibinfo{person}{Paolo Ferragina}, \bibinfo{person}{Dino Pedreschi}, {and} \bibinfo{person}{Fosca Giannotti}.} \bibinfo{year}{2019}\natexlab{}.
\newblock \showarticletitle{A public data set of spatio-temporal match events in soccer competitions}.
\newblock \bibinfo{journal}{\emph{Scientific Data}} \bibinfo{volume}{6}, \bibinfo{number}{1} (\bibinfo{year}{2019}), \bibinfo{pages}{236}.
\newblock


\bibitem[Prasetio et~al\mbox{.}(2016)]%
        {soccer2016ma}
\bibfield{author}{\bibinfo{person}{Darwin Prasetio} {et~al\mbox{.}}} \bibinfo{year}{2016}\natexlab{}.
\newblock \showarticletitle{Predicting football match results with logistic regression}. In \bibinfo{booktitle}{\emph{2016 International Conference On Advanced Informatics: Concepts, Theory And Application}}. IEEE, \bibinfo{pages}{1--5}.
\newblock


\bibitem[Rahman(2020)]%
        {soccer2020rnn}
\bibfield{author}{\bibinfo{person}{Md~Ashiqur Rahman}.} \bibinfo{year}{2020}\natexlab{}.
\newblock \showarticletitle{A deep learning framework for football match prediction}.
\newblock \bibinfo{journal}{\emph{SN Applied Sciences}} \bibinfo{volume}{2}, \bibinfo{number}{2} (\bibinfo{year}{2020}), \bibinfo{pages}{165}.
\newblock


\bibitem[Ramp{\'a}{\v{s}}ek et~al\mbox{.}(2022)]%
        {2022gps}
\bibfield{author}{\bibinfo{person}{Ladislav Ramp{\'a}{\v{s}}ek}, \bibinfo{person}{Michael Galkin}, \bibinfo{person}{Vijay~Prakash Dwivedi}, \bibinfo{person}{Anh~Tuan Luu}, \bibinfo{person}{Guy Wolf}, {and} \bibinfo{person}{Dominique Beaini}.} \bibinfo{year}{2022}\natexlab{}.
\newblock \showarticletitle{Recipe for a general, powerful, scalable graph transformer}.
\newblock \bibinfo{journal}{\emph{Advances in Neural Information Processing Systems}}  \bibinfo{volume}{35} (\bibinfo{year}{2022}), \bibinfo{pages}{14501--14515}.
\newblock


\bibitem[Rong et~al\mbox{.}(2020)]%
        {2020grover}
\bibfield{author}{\bibinfo{person}{Yu Rong}, \bibinfo{person}{Yatao Bian}, \bibinfo{person}{Tingyang Xu}, \bibinfo{person}{Weiyang Xie}, \bibinfo{person}{Ying Wei}, \bibinfo{person}{Wenbing Huang}, {and} \bibinfo{person}{Junzhou Huang}.} \bibinfo{year}{2020}\natexlab{}.
\newblock \showarticletitle{Self-supervised graph transformer on large-scale molecular data}.
\newblock \bibinfo{journal}{\emph{Advances in Neural Information Processing Systems}}  \bibinfo{volume}{33} (\bibinfo{year}{2020}), \bibinfo{pages}{12559--12571}.
\newblock


\bibitem[Rudrapal et~al\mbox{.}(2020)]%
        {soccer2020mlp}
\bibfield{author}{\bibinfo{person}{Dwijen Rudrapal}, \bibinfo{person}{Sasank Boro}, \bibinfo{person}{Jatin Srivastava}, {and} \bibinfo{person}{Shyamu Singh}.} \bibinfo{year}{2020}\natexlab{}.
\newblock \showarticletitle{A deep learning approach to predict football match result}. In \bibinfo{booktitle}{\emph{Computational Intelligence in Data Mining: Proceedings of the International Conference on ICCIDM 2018}}. Springer, \bibinfo{pages}{93--99}.
\newblock


\bibitem[Shazeer et~al\mbox{.}(2017)]%
        {2017moelstm}
\bibfield{author}{\bibinfo{person}{Noam Shazeer}, \bibinfo{person}{Azalia Mirhoseini}, \bibinfo{person}{Krzysztof Maziarz}, \bibinfo{person}{Andy Davis}, \bibinfo{person}{Quoc Le}, \bibinfo{person}{Geoffrey Hinton}, {and} \bibinfo{person}{Jeff Dean}.} \bibinfo{year}{2017}\natexlab{}.
\newblock \showarticletitle{Outrageously large neural networks: The sparsely-gated mixture-of-experts layer}.
\newblock \bibinfo{journal}{\emph{arXiv preprint arXiv:1701.06538}} (\bibinfo{year}{2017}).
\newblock


\bibitem[Shirzad et~al\mbox{.}(2023)]%
        {2023exphormer}
\bibfield{author}{\bibinfo{person}{Hamed Shirzad}, \bibinfo{person}{Ameya Velingker}, \bibinfo{person}{Balaji Venkatachalam}, \bibinfo{person}{Danica~J Sutherland}, {and} \bibinfo{person}{Ali~Kemal Sinop}.} \bibinfo{year}{2023}\natexlab{}.
\newblock \showarticletitle{Exphormer: Sparse transformers for graphs}. In \bibinfo{booktitle}{\emph{International Conference on Machine Learning}}. PMLR, \bibinfo{pages}{31613--31632}.
\newblock


\bibitem[Simpson et~al\mbox{.}(2022)]%
        {2022seq2event}
\bibfield{author}{\bibinfo{person}{Ian Simpson}, \bibinfo{person}{Ryan~J Beal}, \bibinfo{person}{Duncan Locke}, {and} \bibinfo{person}{Timothy~J Norman}.} \bibinfo{year}{2022}\natexlab{}.
\newblock \showarticletitle{Seq2event: Learning the language of soccer using transformer-based match event prediction}. In \bibinfo{booktitle}{\emph{Proceedings of the 28th ACM SIGKDD Conference on Knowledge Discovery and Data Mining}}. \bibinfo{pages}{3898--3908}.
\newblock


\bibitem[Tax and Joustra(2015)]%
        {soccer2015ma}
\bibfield{author}{\bibinfo{person}{Niek Tax} {and} \bibinfo{person}{Yme Joustra}.} \bibinfo{year}{2015}\natexlab{}.
\newblock \showarticletitle{Predicting the Dutch football competition using public data: A machine learning approach}.
\newblock \bibinfo{journal}{\emph{Transactions on Knowledge and Data Engineering}} \bibinfo{volume}{10}, \bibinfo{number}{10} (\bibinfo{year}{2015}), \bibinfo{pages}{1--13}.
\newblock


\bibitem[Vaswani et~al\mbox{.}(2017)]%
        {2017transformer}
\bibfield{author}{\bibinfo{person}{Ashish Vaswani}, \bibinfo{person}{Noam Shazeer}, \bibinfo{person}{Niki Parmar}, \bibinfo{person}{Jakob Uszkoreit}, \bibinfo{person}{Llion Jones}, \bibinfo{person}{Aidan~N Gomez}, \bibinfo{person}{{\L}ukasz Kaiser}, {and} \bibinfo{person}{Illia Polosukhin}.} \bibinfo{year}{2017}\natexlab{}.
\newblock \showarticletitle{Attention is all you need}.
\newblock \bibinfo{journal}{\emph{Advances in Neural Information Processing Systems}}  \bibinfo{volume}{30} (\bibinfo{year}{2017}).
\newblock


\bibitem[Veli{\v{c}}kovi{\'c} et~al\mbox{.}(2017)]%
        {2017gat}
\bibfield{author}{\bibinfo{person}{Petar Veli{\v{c}}kovi{\'c}}, \bibinfo{person}{Guillem Cucurull}, \bibinfo{person}{Arantxa Casanova}, \bibinfo{person}{Adriana Romero}, \bibinfo{person}{Pietro Lio}, {and} \bibinfo{person}{Yoshua Bengio}.} \bibinfo{year}{2017}\natexlab{}.
\newblock \showarticletitle{Graph attention networks}.
\newblock \bibinfo{journal}{\emph{arXiv preprint arXiv:1710.10903}} (\bibinfo{year}{2017}).
\newblock


\bibitem[Wang et~al\mbox{.}(2019)]%
        {2019han}
\bibfield{author}{\bibinfo{person}{Xiao Wang}, \bibinfo{person}{Houye Ji}, \bibinfo{person}{Chuan Shi}, \bibinfo{person}{Bai Wang}, \bibinfo{person}{Yanfang Ye}, \bibinfo{person}{Peng Cui}, {and} \bibinfo{person}{Philip~S Yu}.} \bibinfo{year}{2019}\natexlab{}.
\newblock \showarticletitle{Heterogeneous graph attention network}. In \bibinfo{booktitle}{\emph{The World Wide Web Conference}}. \bibinfo{pages}{2022--2032}.
\newblock


\bibitem[Wu et~al\mbox{.}(2019)]%
        {2019simplifygcn}
\bibfield{author}{\bibinfo{person}{Felix Wu}, \bibinfo{person}{Amauri Souza}, \bibinfo{person}{Tianyi Zhang}, \bibinfo{person}{Christopher Fifty}, \bibinfo{person}{Tao Yu}, {and} \bibinfo{person}{Kilian Weinberger}.} \bibinfo{year}{2019}\natexlab{}.
\newblock \showarticletitle{Simplifying graph convolutional networks}. In \bibinfo{booktitle}{\emph{International Conference on Machine Learning}}. PMLR, \bibinfo{pages}{6861--6871}.
\newblock


\bibitem[Wu et~al\mbox{.}(2024)]%
        {2024sgformer}
\bibfield{author}{\bibinfo{person}{Qitian Wu}, \bibinfo{person}{Wentao Zhao}, \bibinfo{person}{Chenxiao Yang}, \bibinfo{person}{Hengrui Zhang}, \bibinfo{person}{Fan Nie}, \bibinfo{person}{Haitian Jiang}, \bibinfo{person}{Yatao Bian}, {and} \bibinfo{person}{Junchi Yan}.} \bibinfo{year}{2024}\natexlab{}.
\newblock \showarticletitle{Simplifying and empowering transformers for large-graph representations}.
\newblock \bibinfo{journal}{\emph{Advances in Neural Information Processing Systems}}  \bibinfo{volume}{36} (\bibinfo{year}{2024}).
\newblock


\bibitem[Wu et~al\mbox{.}(2021)]%
        {2021graphtrans}
\bibfield{author}{\bibinfo{person}{Zhanghao Wu}, \bibinfo{person}{Paras Jain}, \bibinfo{person}{Matthew Wright}, \bibinfo{person}{Azalia Mirhoseini}, \bibinfo{person}{Joseph~E Gonzalez}, {and} \bibinfo{person}{Ion Stoica}.} \bibinfo{year}{2021}\natexlab{}.
\newblock \showarticletitle{Representing long-range context for graph neural networks with global attention}.
\newblock \bibinfo{journal}{\emph{Advances in Neural Information Processing Systems}}  \bibinfo{volume}{34} (\bibinfo{year}{2021}), \bibinfo{pages}{13266--13279}.
\newblock


\bibitem[Xu et~al\mbox{.}(2018)]%
        {2018gin}
\bibfield{author}{\bibinfo{person}{Keyulu Xu}, \bibinfo{person}{Weihua Hu}, \bibinfo{person}{Jure Leskovec}, {and} \bibinfo{person}{Stefanie Jegelka}.} \bibinfo{year}{2018}\natexlab{}.
\newblock \showarticletitle{How powerful are graph neural networks?}
\newblock \bibinfo{journal}{\emph{arXiv preprint arXiv:1810.00826}} (\bibinfo{year}{2018}).
\newblock


\bibitem[Yu et~al\mbox{.}(2022)]%
        {2022mhgcn}
\bibfield{author}{\bibinfo{person}{Pengyang Yu}, \bibinfo{person}{Chaofan Fu}, \bibinfo{person}{Yanwei Yu}, \bibinfo{person}{Chao Huang}, \bibinfo{person}{Zhongying Zhao}, {and} \bibinfo{person}{Junyu Dong}.} \bibinfo{year}{2022}\natexlab{}.
\newblock \showarticletitle{Multiplex heterogeneous graph convolutional network}. In \bibinfo{booktitle}{\emph{Proceedings of the 28th ACM SIGKDD Conference on Knowledge Discovery and Data Mining}}. \bibinfo{pages}{2377--2387}.
\newblock


\bibitem[Yun et~al\mbox{.}(2019)]%
        {2019gtn}
\bibfield{author}{\bibinfo{person}{Seongjun Yun}, \bibinfo{person}{Minbyul Jeong}, \bibinfo{person}{Raehyun Kim}, \bibinfo{person}{Jaewoo Kang}, {and} \bibinfo{person}{Hyunwoo~J Kim}.} \bibinfo{year}{2019}\natexlab{}.
\newblock \showarticletitle{Graph transformer networks}.
\newblock \bibinfo{journal}{\emph{Advances in Neural Information Processing Systems}}  \bibinfo{volume}{32} (\bibinfo{year}{2019}).
\newblock


\bibitem[Zhang et~al\mbox{.}(2019)]%
        {2019hetergnn}
\bibfield{author}{\bibinfo{person}{Chuxu Zhang}, \bibinfo{person}{Dongjin Song}, \bibinfo{person}{Chao Huang}, \bibinfo{person}{Ananthram Swami}, {and} \bibinfo{person}{Nitesh~V Chawla}.} \bibinfo{year}{2019}\natexlab{}.
\newblock \showarticletitle{Heterogeneous graph neural network}. In \bibinfo{booktitle}{\emph{Proceedings of the 25th ACM SIGKDD International Conference on Knowledge Discovery \& Data Mining}}. \bibinfo{pages}{793--803}.
\newblock


\bibitem[Zhang et~al\mbox{.}(2018)]%
        {2018networksurvey}
\bibfield{author}{\bibinfo{person}{Daokun Zhang}, \bibinfo{person}{Jie Yin}, \bibinfo{person}{Xingquan Zhu}, {and} \bibinfo{person}{Chengqi Zhang}.} \bibinfo{year}{2018}\natexlab{}.
\newblock \showarticletitle{Network representation learning: A survey}.
\newblock \bibinfo{journal}{\emph{IEEE Transactions on Big Data}} \bibinfo{volume}{6}, \bibinfo{number}{1} (\bibinfo{year}{2018}), \bibinfo{pages}{3--28}.
\newblock


\bibitem[Zhang et~al\mbox{.}(2020)]%
        {2020graphbert}
\bibfield{author}{\bibinfo{person}{Jiawei Zhang}, \bibinfo{person}{Haopeng Zhang}, \bibinfo{person}{Congying Xia}, {and} \bibinfo{person}{Li Sun}.} \bibinfo{year}{2020}\natexlab{}.
\newblock \showarticletitle{Graph-bert: Only attention is needed for learning graph representations}.
\newblock \bibinfo{journal}{\emph{arXiv preprint arXiv:2001.05140}} (\bibinfo{year}{2020}).
\newblock


\bibitem[Zhao et~al\mbox{.}(2022)]%
        {2022winningtracker}
\bibfield{author}{\bibinfo{person}{Chuang Zhao}, \bibinfo{person}{Hongke Zhao}, \bibinfo{person}{Yong Ge}, \bibinfo{person}{Runze Wu}, {and} \bibinfo{person}{Xudong Shen}.} \bibinfo{year}{2022}\natexlab{}.
\newblock \showarticletitle{Winning tracker: a new model for real-time winning prediction in MOBA games}. In \bibinfo{booktitle}{\emph{Proceedings of the ACM Web Conference 2022}}. \bibinfo{pages}{3387--3395}.
\newblock


\bibitem[Zhao et~al\mbox{.}(2023)]%
        {zhao2023teamgraph}
\bibfield{author}{\bibinfo{person}{Kai Zhao}, \bibinfo{person}{Chunjie Du}, {and} \bibinfo{person}{Guangxin Tan}.} \bibinfo{year}{2023}\natexlab{}.
\newblock \showarticletitle{Enhancing basketball game outcome prediction through fused graph convolutional networks and random forest algorithm}.
\newblock \bibinfo{journal}{\emph{Entropy}} \bibinfo{volume}{25}, \bibinfo{number}{5} (\bibinfo{year}{2023}), \bibinfo{pages}{765}.
\newblock


\bibitem[Zhuang and Ma(2018)]%
        {2018dualgcn}
\bibfield{author}{\bibinfo{person}{Chenyi Zhuang} {and} \bibinfo{person}{Qiang Ma}.} \bibinfo{year}{2018}\natexlab{}.
\newblock \showarticletitle{Dual graph convolutional networks for graph-based semi-supervised classification}. In \bibinfo{booktitle}{\emph{Proceedings of the 2018 World Wide Web Conference}}. \bibinfo{pages}{499--508}.
\newblock


\end{thebibliography}
